\documentclass{article}

\usepackage[pdftex]{graphicx, color}
\usepackage{cite}
\usepackage{amsmath,amssymb,amsfonts}
\usepackage{algorithmic}
\usepackage{textcomp}
\usepackage{xcolor}
\usepackage{bm}
\usepackage{multirow}
\usepackage[colorlinks=true, linkcolor=blue, filecolor=blue, urlcolor=blue, citecolor=blue]{hyperref}

\newcommand{\argmax}{\mathop{\rm arg~max}\limits}
\newcommand{\argmin}{\mathop{\rm arg~min}\limits}

\usepackage{authblk}
\usepackage{indentfirst}
\usepackage{geometry}
\allowdisplaybreaks[2]

\makeatletter
\long\def\@makecaption#1#2{%
  \normalsize
  \vskip\abovecaptionskip
  \sbox\@tempboxa{#1: #2}%
  \ifdim \wd\@tempboxa >\hsize
    #1: #2\par
  \else
    \global \@minipagefalse
    \hb@xt@\hsize{\hfil\box\@tempboxa\hfil}%
  \fi
  \vskip\belowcaptionskip}
\makeatother

\title{AutoLL: Automatic Linear Layout of Graphs based on Deep Neural Network}
\author[1]{Chihiro Watanabe\thanks{watanabe-chihiro763@g.ecc.u-tokyo.ac.jp}} 
\author[1,2]{Taiji Suzuki\thanks{taiji@mist.i.u-tokyo.ac.jp}}
\affil[1]{{\normalsize Graduate School of Information Science and Technology, The University of Tokyo, Tokyo, Japan}}
\affil[2]{{\normalsize Center for Advanced Intelligence Project (AIP), RIKEN, Tokyo, Japan}}
\date{}

\begin{document}
\maketitle

\begin{abstract}
Linear layouts are a graph visualization method that can be used to capture an entry pattern in an adjacency matrix of a given graph. By reordering the node indices of the original adjacency matrix, linear layouts provide knowledge of latent graph structures. Conventional linear layout methods commonly aim to find an optimal reordering solution based on predefined features of a given matrix and loss function. However, prior knowledge of the appropriate features to use or structural patterns in a given adjacency matrix is not always available. In such a case, performing the reordering based on data-driven feature extraction without assuming a specific structure in an adjacency matrix is preferable. 
Recently, a neural-network-based matrix reordering method called DeepTMR has been proposed to perform this function. However, it is limited to a two-mode reordering (i.e., the rows and columns are reordered separately) and it cannot be applied in the one-mode setting (i.e., the same node order is used for reordering both rows and columns), owing to the characteristics of its model architecture. In this study, we extend DeepTMR and propose a new one-mode linear layout method referred to as AutoLL. We developed two types of neural network models, AutoLL-D and AutoLL-U, for reordering directed and undirected networks, respectively. To perform one-mode reordering, these AutoLL models have specific encoder architectures, which extract node features from an observed adjacency matrix. 
We conducted both qualitative and quantitative evaluations of the proposed approach, and the experimental results demonstrate its effectiveness. 

\end{abstract}

\section{Introduction}
\label{sec:introduction}

\begin{figure}[tp]
\centerline{\includegraphics[width=0.6\hsize]{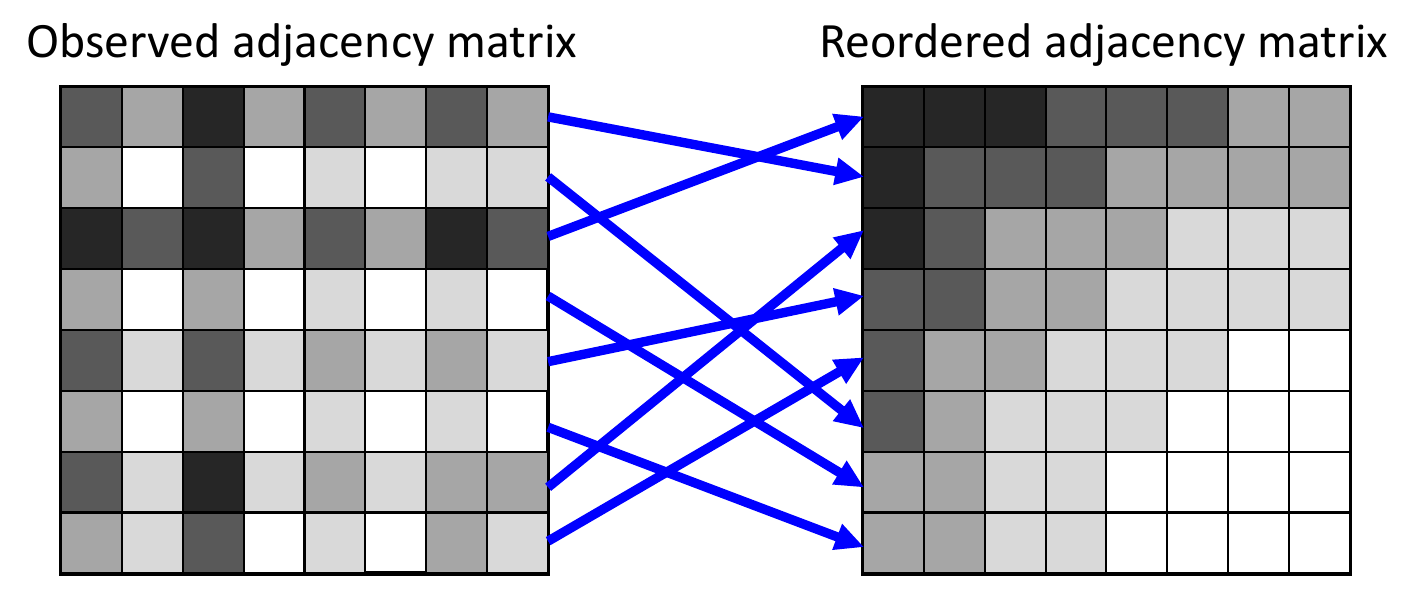}}
\caption{Graph reordering problem. We assume that some structural patterns appear in a given observed adjacency matrix by reordering the node indices.}
\label{fig:gr}
\end{figure}

Linear graph layouts, also known as matrix reordering or seriation \cite{Diaz2002, Behrisch2016, Liiv2010}, are a matrix visualization technique that has been used to analyze various data matrices such as archaeological and sociological data \cite{Ihm2005, Forsyth1946}. 
The goal of linear layouts is to find orders (i.e., permutations) of row and column indices of a given matrix such that the reordered matrix shows some structural pattern (Fig.~\ref{fig:gr}). Specifically, we focus on the one-mode reordering setting and call this task \textit{graph reordering}, where the same order of node indices is applied to \textit{both} rows and columns of the adjacency matrix $A$ of a given (weighted or unweighted) graph, regardless of whether the graph is directed (i.e., whether $A$ is symmetric) \cite{Hahsler2017, Imaizumi2020}. 

Various methods have been proposed to solve such linear layout problems. For instance, among the seven categories of linear layout methods introduced by \cite{Behrisch2016}, the Robinsonian and graph-theoretic approaches seek an optimal node reordering with regard to predefined loss functions such as bandwidth and profile \cite{Robinson1951, Leung1984, Lin1994, Zhao2020}. However, computational cost is a notable problem for these combinatorial-optimization-based methods. Obtaining a globally optimal solution for a given loss function from all possible reordering patterns becomes intractable with increasing matrix size. Several spectral/dimension-reduction methods have been proposed as alternatives, which assume that an observed matrix $A$ has some specific low-dimensional structure \cite{Friendly2002, Liu2003, Rodgers1992, Spence1974}. Based on this assumption, we apply a low-dimensional approximation to matrix $A$ or some other feature matrix defined by using matrix $A$, and determine the node order based on the approximation results. Such a low-dimensional assumption brings several advantages, in that it obviates the necessity of solving any combinatorial optimization problem directly, and the reordering is less sensitive to the noise in matrix entries. 

Although the specific algorithms of the conventional linear layout methods differ, most follow the same common process of first computing a predefined feature from a given observed matrix and then finding the (approximately) optimal reordering solution based on such a feature and a given loss function. In practice, however, prior knowledge as to the appropriate features to use or structural patterns in a given adjacency matrix is not always available. In such cases, it would be preferable to avoid the requirement of specifying the feature extraction function or expected structure in a given adjacency matrix beforehand. 

Recently, a neural-network-based matrix reordering method called DeepTMR \cite{Watanabe2021} has been proposed for two-mode matrix data (i.e., the rows and columns generally represent different objects) to fulfill the abovementioned objective. Although it cannot be applied to one-mode reordering (i.e., the rows and columns represent the same object and therefore the same order is applied to rows and columns), this approach has several advantages. First, using the above autoencoder-like model avoids the necessity of specifying a latent structural pattern in a given matrix $A$ or defining the features for row/column reordering in advance. Moreover, the output of the DeepTMR model can be seen as a denoised mean matrix, which provides us with knowledge about the global structural pattern in the original matrix $A$. These advantages are realized by using a model with an autoencoder-like architecture, where two encoders extract row and column features from a given observed matrix $A$ and a decoder reconstructs each entry of matrix $A$ from these features. After training the entire network to minimize the reconstruction error, the rows and columns can be reordered based on the extracted features of the trained encoders. 

In this paper, we extend DeepTMR and propose a new neural-network-based graph reordering method called AutoLL. The main difference between the proposed AutoLL and DeepTMR is that the former can be used for reordering one-mode adjacency matrices, whereas the latter can be used for two-mode relational data matrices. Because DeepTMR consists of two separate encoders for extracting row and column features, it cannot be applied to our one-mode setting. To solve this problem, we adopt specific encoder architectures to enable two trained models, AutoLL-D and AutoLL-U, to be used for reordering directed and undirected networks, respectively. From observed adjacency matrix $A$, these encoders extract one-dimensional node features, which can be used for reordering \textit{both} rows and columns of matrix $A$. 

The remainder of this study is organized as follows. First, we explain the proposed AutoLL in Section \ref{sec:method}. In Section \ref{sec:experiments}, we conduct experiments on both qualitative and quantitative evaluations of the proposed AutoLL using synthetic and practical datasets. We discuss the features and limitations of AutoLL in Section \ref{sec:discussion}, and conclude this paper in Section \ref{sec:conclusion}. 


\section{AutoLL: automatic linear layout of graphs based on deep neural network}
\label{sec:method}

\begin{figure}[tp]
\centerline{\includegraphics[width=0.8\hsize]{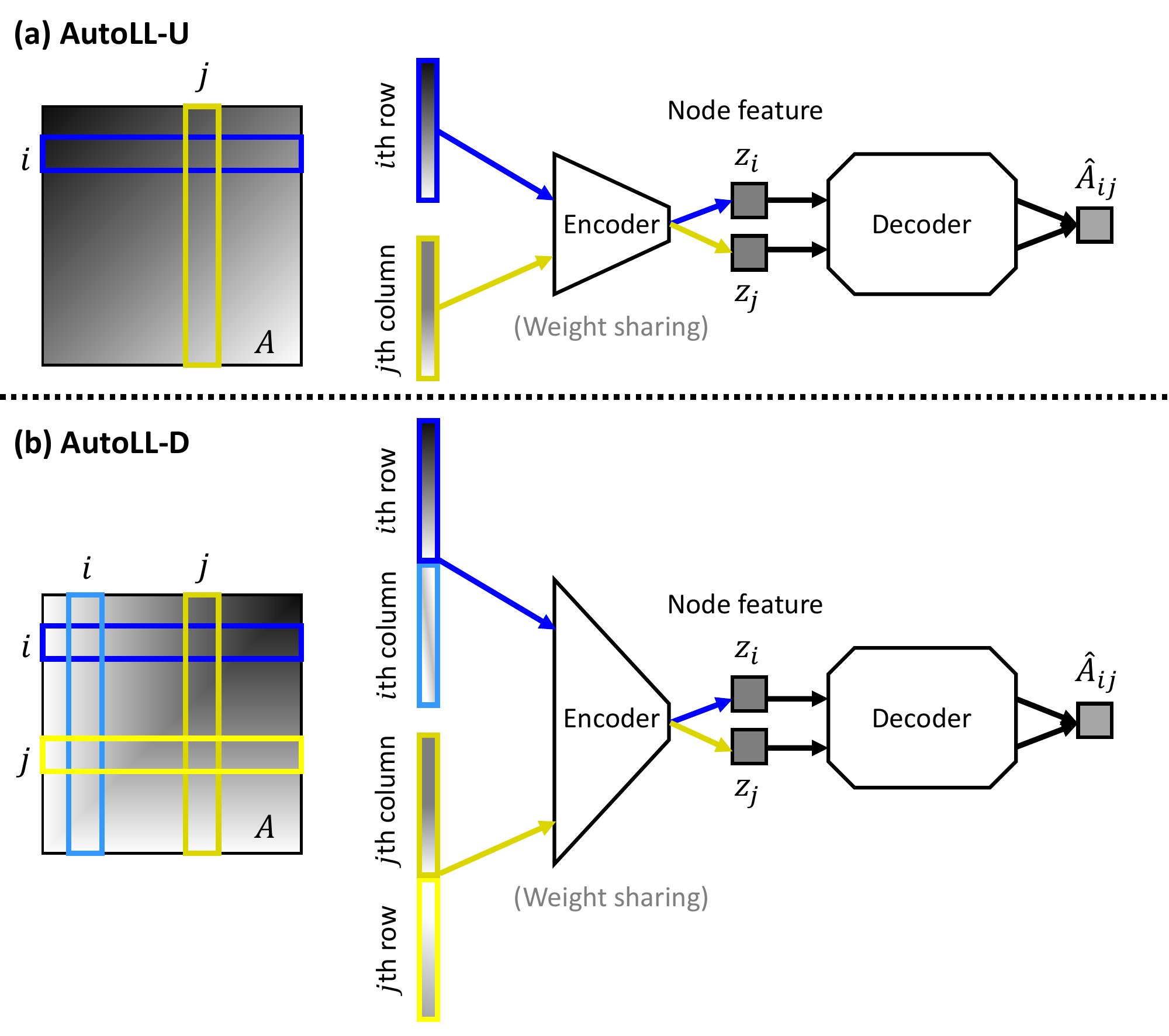}}
\caption{Network architecture of the proposed AutoLL models. (a) AutoLL-U and (b) AutoLL-D for undirected and directed graphs, respectively.}
\label{fig:architecture}
\end{figure}

We propose a new neural-network-based graph reordering method called AutoLL. Fig.~\ref{fig:architecture} shows the entire network architectures of both the proposed models AutoLL-U and AutoLL-D, which can be applied to undirected and directed networks, respectively. Our aim is to determine the order of the node indices $\{ 1, 2, \dots, n \}$ in a given graph with adjacency matrix $A = (A_{ij})_{1 \leq i, j \leq n} \in \mathbb{R}^{n \times n}$ such that the reordered adjacency matrix $\underline{A}$ shows some structural pattern. As explained in Section \ref{sec:introduction}, we consider a one-mode reordering setting in which the same node order is applied to both rows and columns of matrix $A$. We describe the graph reordering procedure used by the AutoLL-U and AutoLL-D models below. 

\paragraph{AutoLL-U: reordering of undirected graphs} For undirected graphs, we construct an autoencoder-like model based on the assumption that each entry $A_{ij}$ of a given adjacency matrix $A$ can be reconstructed from the $i$th row $\bm{r}^{(i)}$ and the $j$th column $\bm{c}^{(j)}$ of matrix $A$, which are given by
\begin{align}
\label{eq:row_col_vectors}
\bm{r}^{(i)} = (r^{(i)}_{j'})_{1 \leq j' \leq n}, \ \ \ r^{(i)}_{j'} = A_{ij'}, 
\nonumber \\
\bm{c}^{(j)} = (c^{(j)}_{i'})_{1 \leq i' \leq n}, \ \ \ c^{(j)}_{i'} = A_{i'j}. 
\end{align}
It must be noted that $\bm{r}^{(i)} = \bm{c}^{(i)}$ holds for all $i \in \{ 1, \dots, n \}$, because matrix $A$ is symmetric. Based on this assumption, the AutoLL-U model takes $(\bm{r}^{(i)}, \bm{c}^{(j)})$ and $A_{ij}$ as a pair of input and output data points, respectively. From these input row and column vectors, the encoder part of the AutoLL-U model extracts one-dimensional node features, which are later used to determine the order of the node indices. Because we consider a case in which matrix $A$ is symmetric, to conduct one-mode matrix reordering, a single encoder is shared for both row and column input vectors. 
\begin{align}
\label{feature_z_u}
z_i = \mathrm{ENC} \left( \bm{r}^{(i)} \right), \ \ \ \ \ 
z_j = \mathrm{ENC} \left( \bm{c}^{(j)} \right), 
\end{align}
where $\mathrm{ENC}: \mathbb{R}^n \mapsto \mathbb{R}$ is an arbitrary encoder function. Specifically, we assume that this encoder function is represented as a neural network trained with a given data matrix. 
Based on these row and column features, the decoder part of AutoLL-U reconstructs the $(i, j)$th entry of matrix $A$ as follows. 
\begin{align}
\hat{A}_{ij} = \mathrm{DEC} (z_i, z_j), 
\end{align}
where $\mathrm{DEC}: \mathbb{R}^2 \mapsto \mathbb{R}$ is an arbitrary decoder function. Along with the encoder function, this decoder function is trained as a neural network with a given data matrix. 

\paragraph{AutoLL-D: reordering of directed graphs} The above AutoLL-U model assumes that the $i$th row and column vectors are identical (i.e., $\bm{r}^{(i)} = \bm{c}^{(i)}$). For directed graphs, this condition does not hold; therefore, we construct another model based on the assumption that each entry $A_{ij}$ of a given adjacency matrix $A$ can be reconstructed from the information of the $i$th and $j$th row and column vectors $(\bm{r}^{(i)}, \bm{c}^{(i)}, \bm{r}^{(j)} \bm{c}^{(j)})$ of matrix $A$, where $\{\bm{r}^{(i)}\}$ and $\{\bm{c}^{(i)}\}$ are given by \eqref{eq:row_col_vectors}. Based on this assumption, the AutoLL-D model takes $(\bm{v}^{(i)}, \bm{v}^{(j)})$ and $A_{ij}$ as a pair of input and output data points, where
\begin{align}
\label{eq:feature_v}
\bm{v}^{(i)} = \begin{bmatrix}
\left(\bm{r}^{(i)}\right)^{\top} \left(\bm{c}^{(i)}\right)^{\top}
\end{bmatrix}^{\top}, \ \ \ i = 1, \dots, n. 
\end{align}
Of note, if $i=j$, then $\bm{v}^{(i)} = \bm{v}^{(j)}$ holds from the definition in \eqref{eq:feature_v}. As in AutoLL-U, the shallower part of the AutoLL-D model extracts one-dimensional node features from these row and column vectors with a single encoder. 
\begin{align}
\label{feature_z_d}
z_i = \mathrm{ENC} \left( \bm{v}^{(i)} \right), \ \ \ \ \ 
z_j = \mathrm{ENC} \left( \bm{v}^{(j)} \right), 
\end{align}
where $\mathrm{ENC}: \mathbb{R}^n \mapsto \mathbb{R}$ is an arbitrary encoder function. 
Based on these features, the $(i, j)$th entry of the matrix $A$ is estimated as
\begin{align}
\hat{A}_{ij} = \mathrm{DEC} (z_i, z_j), 
\end{align}
where $\mathrm{DEC}: \mathbb{R}^2 \mapsto \mathbb{R}$ is an arbitrary decoder function. 

Both the AutoLL-U and AutoLL-D models are trained to minimize the following reconstruction error with the $L_2$ regularization term. 
\begin{align}
\label{eq:loss}
\mathcal{L} = \frac{1}{|\mathcal{I}_t|} \sum_{(i, j) \in \mathcal{I}_t} \mathrm{BCE} (\hat{A}_{ij}, A_{ij}) + \lambda \| \bm{w} \|_2^2, 
\end{align}
where $\mathrm{BCE} (\cdot, \cdot)$ is the binary cross-entropy (i.e., $\mathrm{BCE} (x, y) = -[y \log x + (1-y) \log (1-x)]$), $\mathcal{I}_t$ is a set of entries $(i, j)$ in the $t$th mini-batch, $\lambda$ is a hyperparameter, and $\bm{w}$ is a vector of all the learnable model parameters. In the experiments in Section \ref{sec:experiments}, we normalized the observed adjacency matrix $A$ such that its maximum and minimum entry values were one and zero, respectively. Moreover, by using a sigmoid activation function in the last layer of the decoder, we set the interval of the model output $\hat{A}_{ij}$ to $(0, 1)$. Based on these procedures, the above binary cross-entropy was defined for any observed adjacency matrix $A \in \mathbb{R}^{n \times n}$. 

By using the trained AutoLL models, we can finally reorder the node indices of a given graph. For each $i$th node of matrix $A$, we define the input vector as in \eqref{eq:row_col_vectors} and \eqref{eq:feature_v}, and compute the node feature $\bm{z} = (z_i)_{1 \leq i \leq n}$ as in (\ref{feature_z_u}) and (\ref{feature_z_d}). Because the decoder is trained to reconstruct each $(i, j)$th entry of matrix $A$ only from $z_i$ and $z_j$, we can expect that these node features reflect the global structural pattern in matrix $A$. Therefore, the proposed approach reorders the node indices based on this feature. Specifically, we define the order $\pi$ of the nodes as a permutation of $\{ 1, 2, \dots, n \}$ such that $z_{\pi (1)} \leq z_{\pi (2)} \leq \dots \leq z_{\pi (n)}$ holds. Based on this definition, the reordered versions of the observed and reconstructed adjacency matrices and the node feature, $\underline{A}$, $\underline{\hat{A}}$, and $\underline{\bm{z}}$, respectively, are given by
\begin{align}
&\underline{A} = (\underline{A}_{ij})_{1 \leq i, j \leq n}, \ \ \ \underline{A}_{ij} = A_{\pi (i) \pi (j)}, \ \ \ i, j = 1, \dots, n, \nonumber \\
&\underline{\hat{A}} = (\underline{\hat{A}}_{ij})_{1 \leq i, j \leq n}, \ \ \ \underline{\hat{A}}_{ij} = \hat{A}_{\pi (i) \pi (j)}, \ \ \ i, j = 1, \dots, n, \nonumber \\
&\underline{\bm{z}} = (\underline{z}_i)_{1 \leq i \leq n}, \ \ \ \underline{z}_i = z_{\pi (i)}, \ \ \ i = 1, \dots, n. 
\end{align}


\section{Experiments}
\label{sec:experiments}

We experimentally demonstrate the effectiveness of the proposed AutoLL models through both qualitative and quantitative evaluations. For all the experiments, we used the following settings. 
\begin{itemize}
\item We used the parameter initialization method in \cite{Glorot2010} for the weights of the linear layers, where the initial weight between the $l$th and $(l+1)$th layers was generated from a uniform distribution on interval $\left[ -1/\sqrt{m^{(l)}}, 1/\sqrt{m^{(l)}} \right]$. Here, we denote the number of units in the $l$th layer as $m^{(l)}$. We set the initial biases to zero. 
\item We used the Adam optimizer \cite{Kingma2015} to train the AutoLL models, with $\beta_1 = 0.9$, $\beta_2 = 0.999$, $\epsilon = 1.0 \times 10^{-8}$, and the learning rate $\eta = 1.0 \times 10^{-2}$. The regularization hyperparameter was set to $\lambda = 1.0 \times 10^{-10}$. 
\item Table \ref{tab:hyperparameter} shows the hyperparameter settings of the number of epochs $T$ (the total number of iterations is $\mathrm{ceil} [Tn^2/|\mathcal{I}|]$), mini-batch size $|\mathcal{I}|$, and number of units in each layer of the encoder and decoder, $\bm{m}^{\mathrm{ENC}}$ and $\bm{m}^{\mathrm{DEC}}$, respectively (from input to output). Each layer of the AutoLL model is a linear layer followed by a sigmoid activation function. 
\item In all the experiments in Sections \ref{sec:exp_syn} and \ref{sec:exp_compare}, we set the node size at $n = 120$. 
\end{itemize}

\begin{table}[t]
\centering
\caption{Hyperparameter settings in training the AutoLL models. ``-D'' and ``-U'' stand for directed and undirected cases, respectively.}\vspace{3mm}
\begin{tabular}{|c||c|c|c|c|} \hline
& $T$ & $|\mathcal{I}|$ & $\bm{m}^{\mathrm{ENC}}$ & $\bm{m}^{\mathrm{DEC}}$ \\ \hline \hline
Sec. \ref{sec:exp_syn}, SBM-U & \multirow{6}{*}{$2 \times 10^2$} & \multirow{6}{*}{$2 \times 10^2$} & \multirow{3}{*}{$\begin{bmatrix} n, 10, 1 \end{bmatrix}$} & \multirow{8}{*}{$\begin{bmatrix} 2, 10, 1 \end{bmatrix}$} \\ \cline{1-1} 
Sec. \ref{sec:exp_syn}, DGM-U & & & & \\ \cline{1-1}
Sec. \ref{sec:exp_compare}, DGM-U & & & & \\ \cline{1-1} \cline{4-4} 
Sec. \ref{sec:exp_syn}, SBM-D & & & \multirow{3}{*}{$\begin{bmatrix} 2n, 10, 1 \end{bmatrix}$} & \\ \cline{1-1} 
Sec. \ref{sec:exp_syn}, DGM-D & & & & \\ \cline{1-1} 
Sec. \ref{sec:exp_compare}, DGM-D & & & & \\ \cline{1-3} \cline{4-4} 
Sec. \ref{sec:exp_practical} Football network & $1 \times 10^4$ & $5 \times 10^3$ & $\begin{bmatrix} n, 10, 1 \end{bmatrix}$ & \\ \cline{1-4} 
Sec. \ref{sec:exp_practical} Neural network & $5 \times 10^3$ & $3 \times 10^3$ & $\begin{bmatrix} 2n, 10, 1 \end{bmatrix}$ & \\ \hline
\end{tabular}
\label{tab:hyperparameter}
\end{table}

\subsection{Experiments using synthetic datasets based on stochastic block model and diagonal gradation model}
\label{sec:exp_syn}

First, we applied the AutoLL models to synthetic datasets with latent structural patterns and checked the reordering results. In this experiment, we used the following two generative models of an adjacency matrix. 

\paragraph{Stochastic block model (SBM)} In SBM, we assume that each entry of an adjacency matrix is generated independently from a block-wise identical distribution, given a cluster assignment of all the nodes \cite{Holland1983}. 
Specifically, we set the cluster number at $K = 3$ and define the cluster assignment $c_i$ of each $i$th node as $c_{(k - 1) n^{(0)}+1} = \dots = c_{k n^{(0)}} = k$ for $k = 1, \dots, K$, where $n^{(0)} = \frac{n}{K}$. Each entry $\bar{A}^{(0)}_{ij}$ of matrix $\bar{A}^{(0)} \in \mathbb{R}^{n \times n}$ is generated independently from a Gaussian distribution with mean $B_{c_i c_j}$ and standard deviation $\sigma$, where
\begin{align}
B = \begin{bmatrix}
0.9 & 0.1 & 0.3\\
0.4 & 0.8 & 0.2\\
0.1 & 0.3 & 0.7
\end{bmatrix}, \ \ \ 
\sigma = 0.05. 
\end{align}

\paragraph{Diagonal gradation model (DGM)} In DGM, we assume a gradation pattern in an adjacency matrix. Specifically, we generate each entry $\bar{A}^{(0)}_{ij}$ of matrix $\bar{A}^{(0)} \in \mathbb{R}^{n \times n}$ independently from a Gaussian distribution with mean $B_{ij}$ and standard deviation $\sigma$, where
\begin{align}
&B_{ij} = 0.9 - 0.8 \frac{n - 1 - i + j}{2n - 2},\ \ \ i, j = 1, \dots, n, \nonumber \\
&\sigma = 0.05. 
\end{align}

For both models, to evaluate the AutoLL-D and AutoLL-U models, we used matrices $\bar{A}^{(1)} = \bar{A}^{(0)}$ and
\begin{align}
\bar{A}^{(1)} = (\bar{A}^{(1)}_{ij})_{1 \leq i, j \leq n}, \ \ \ 
\bar{A}^{(1)}_{ij} = \begin{cases}
\bar{A}^{(0)}_{ij} & \mathrm{if}\ i \leq j, \\
\bar{A}^{(0)}_{ji} & \mathrm{otherwise}, 
\end{cases}
\end{align}
respectively. As explained in Section \ref{sec:method}, we normalize matrix $\bar{A}^{(1)}$ such that all the entries of the resulting matrix $\bar{A}$ are within the interval of $[0, 1]$, as formulated below. 
\begin{align}
\label{eq:A_normalization}
&\bar{A} = (\bar{A}_{ij})_{1 \leq i, j \leq n}, \nonumber \\
&\bar{A}_{ij} = \frac{\bar{A}^{(1)}_{ij} - \min_{(i, j) \in \mathcal{J}} \bar{A}^{(1)}_{ij}}{\max_{(i, j) \in \mathcal{J}} \bar{A}^{(1)}_{ij} - \min_{(i, j) \in \mathcal{J}} \bar{A}^{(1)}_{ij}}. 
\end{align}
where $\mathcal{J} = \{ (i, j) | i, j = 1, \dots, n \}$. Finally, we defined the observed matrix $A$ by randomly permuting the node indices of matrix $\bar{A}$ and applied the AutoLL model to matrix $A$. 

\begin{figure}[t]
\centerline{\includegraphics[width=0.24\hsize]{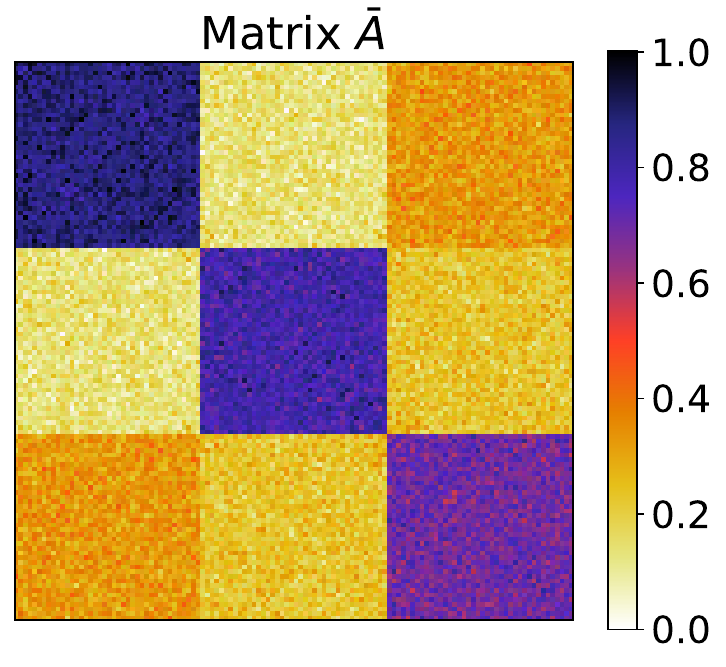}
\includegraphics[width=0.24\hsize]{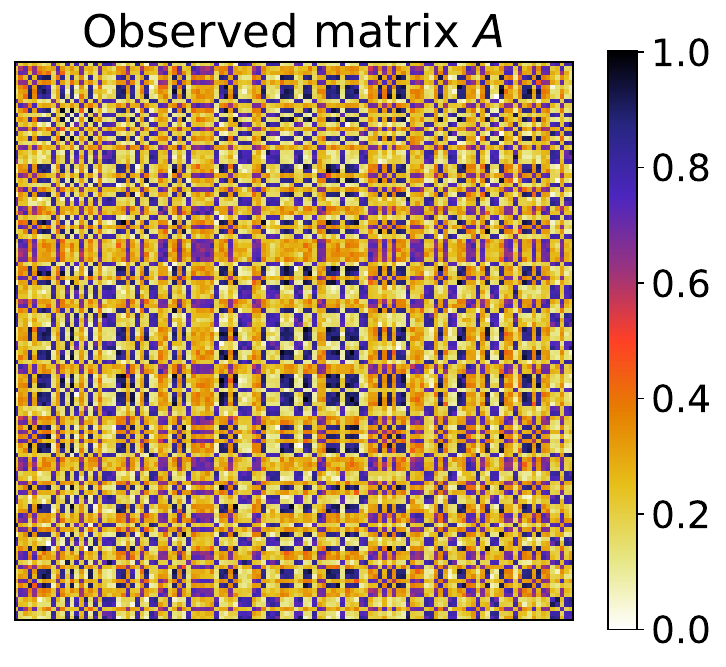}
\includegraphics[width=0.24\hsize]{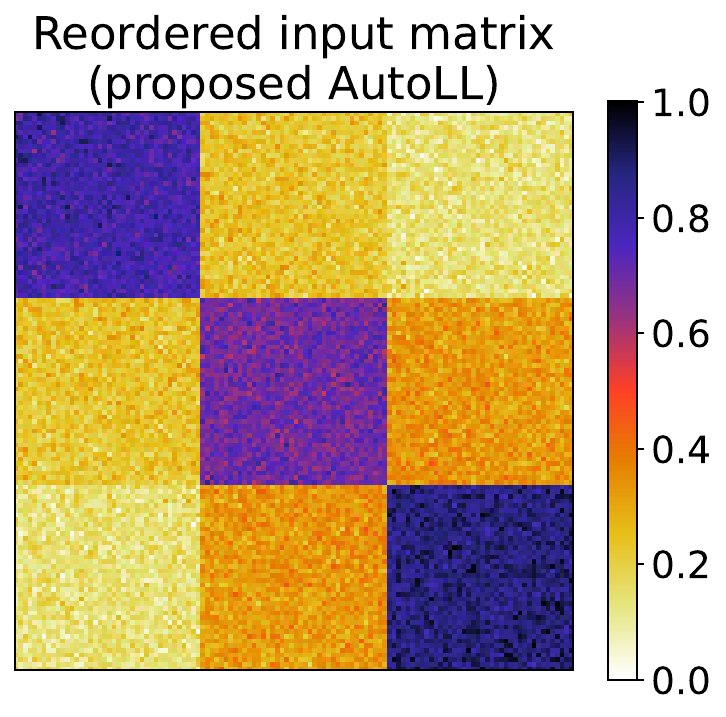}
\includegraphics[width=0.24\hsize]{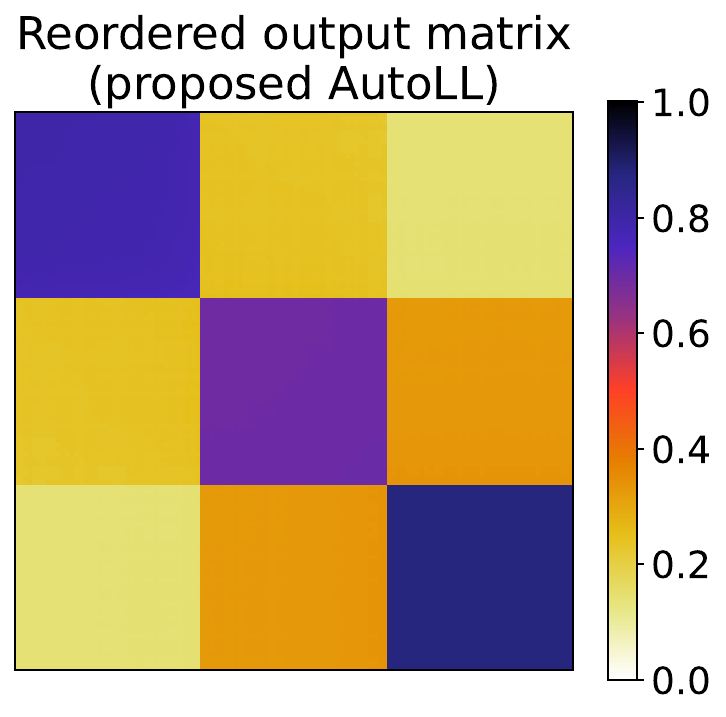}}
\centerline{\includegraphics[width=0.48\hsize]{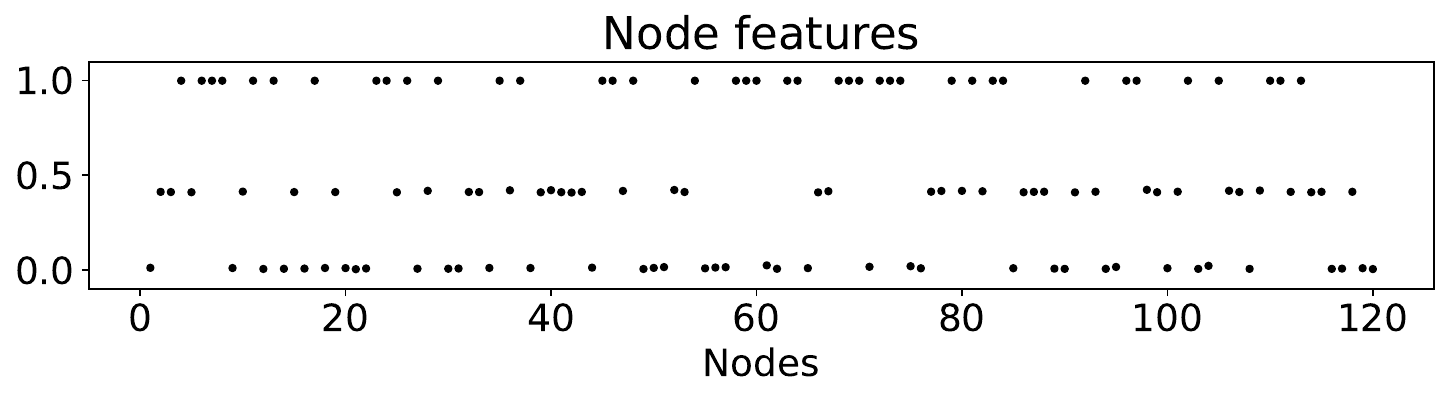}
\includegraphics[width=0.48\hsize]{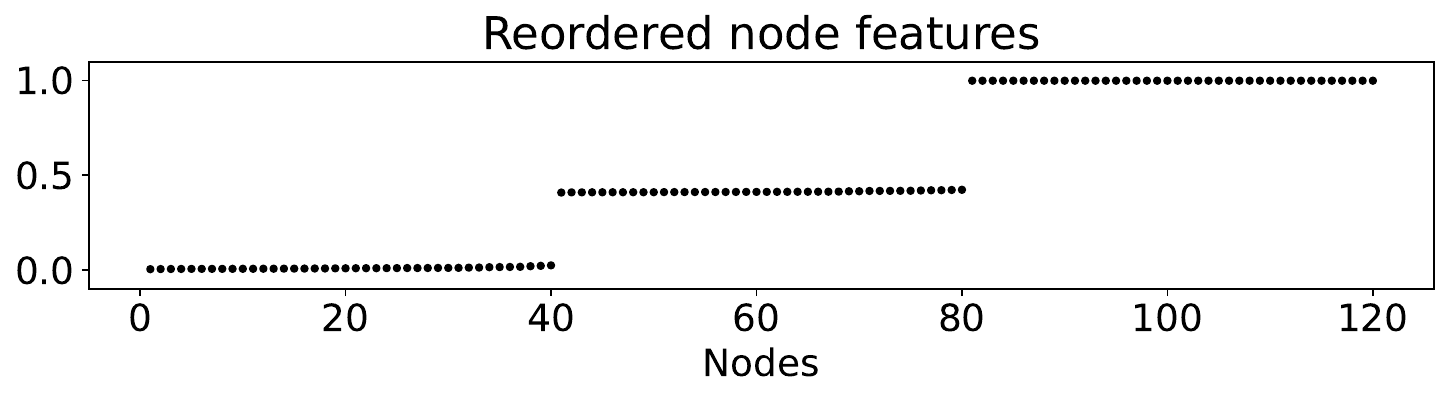}}
\caption{Matrices $\bar{A}$, $A$, $\underline{A}$, and $\underline{\hat{A}}$ and vectors $\bm{z}$ and $\underline{\bm{z}}$ (\textbf{undirected SBM}).}\vspace{3mm}
\label{fig:syn1u}
\centerline{\includegraphics[width=0.24\hsize]{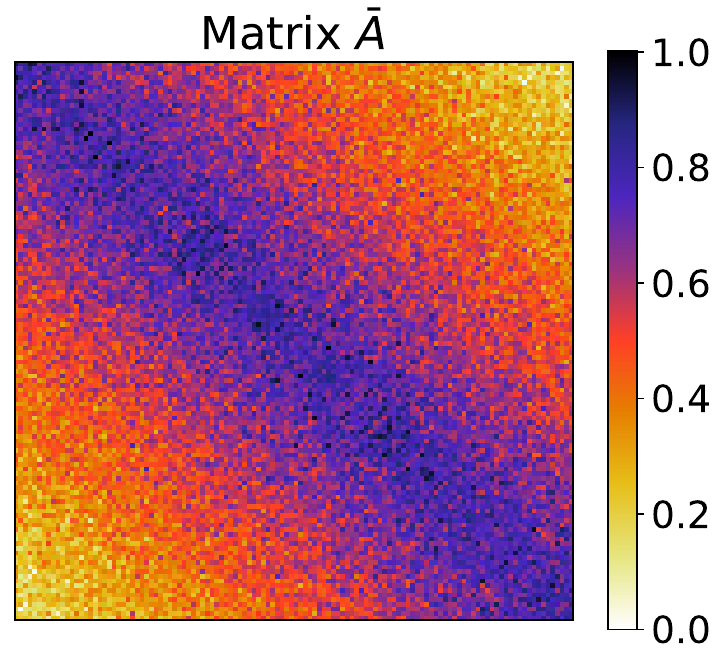}
\includegraphics[width=0.24\hsize]{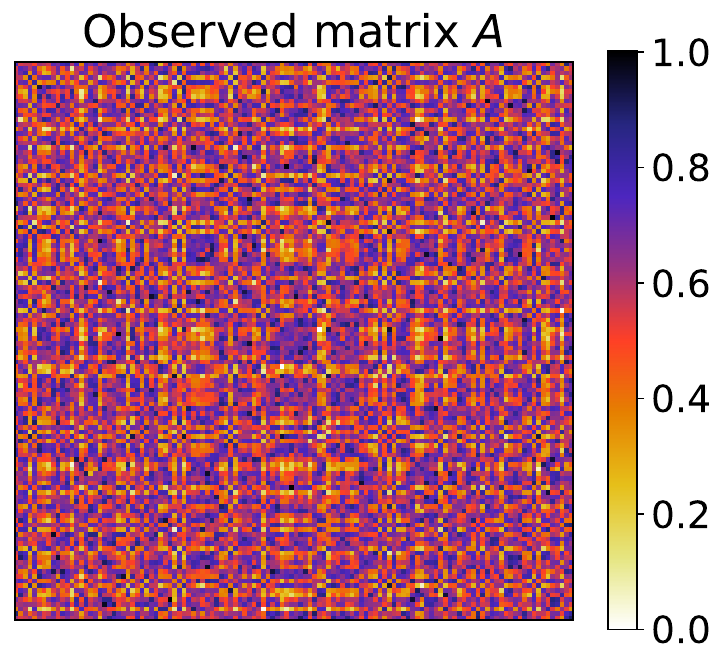}
\includegraphics[width=0.24\hsize]{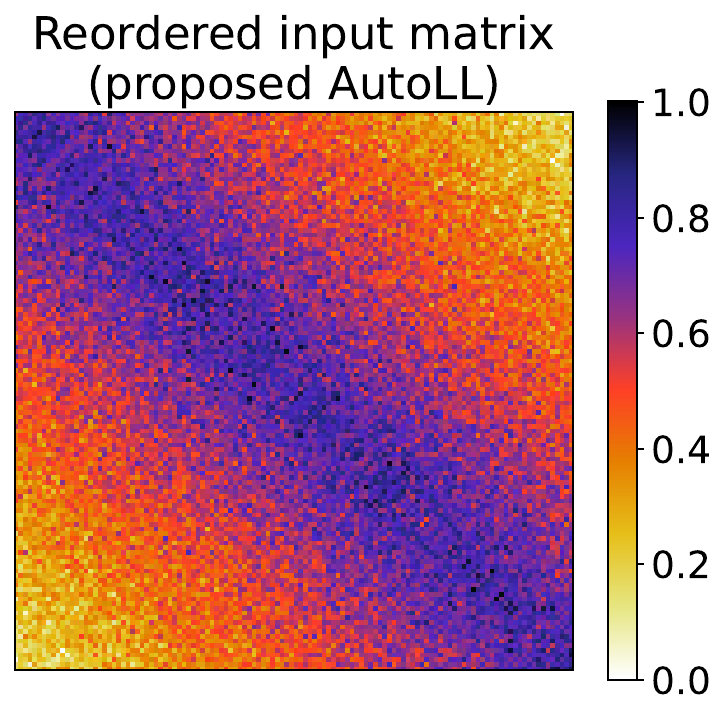}
\includegraphics[width=0.24\hsize]{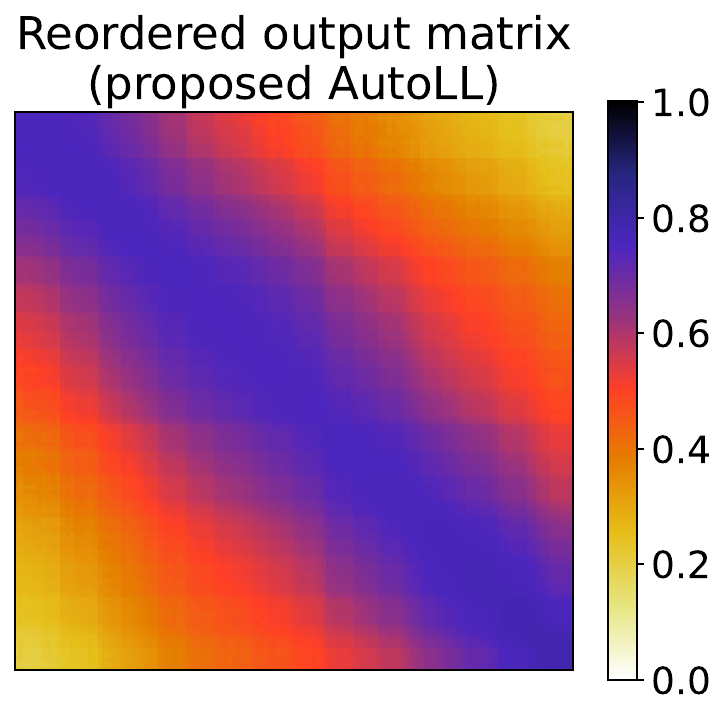}}
\centerline{\includegraphics[width=0.48\hsize]{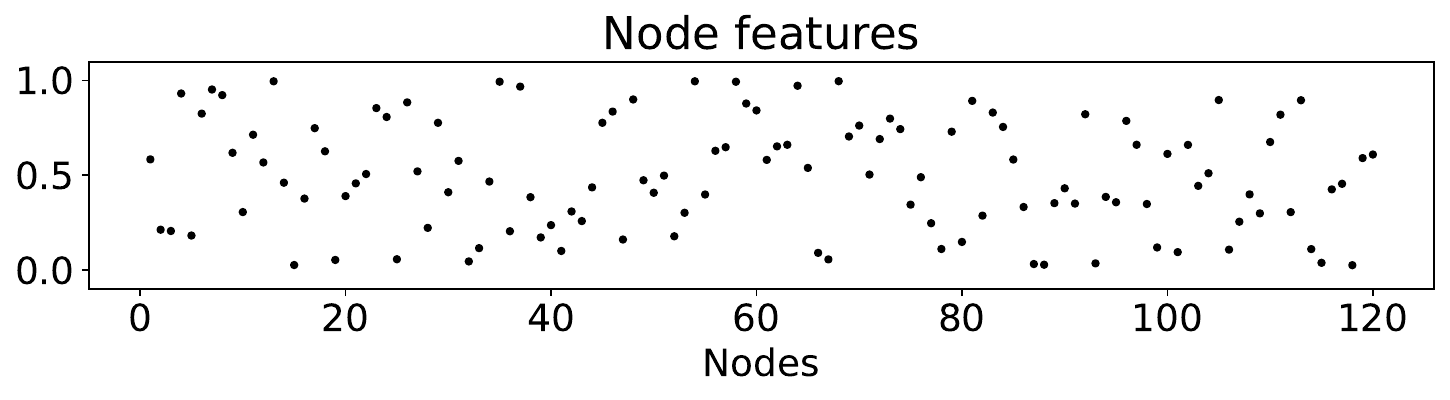}
\includegraphics[width=0.48\hsize]{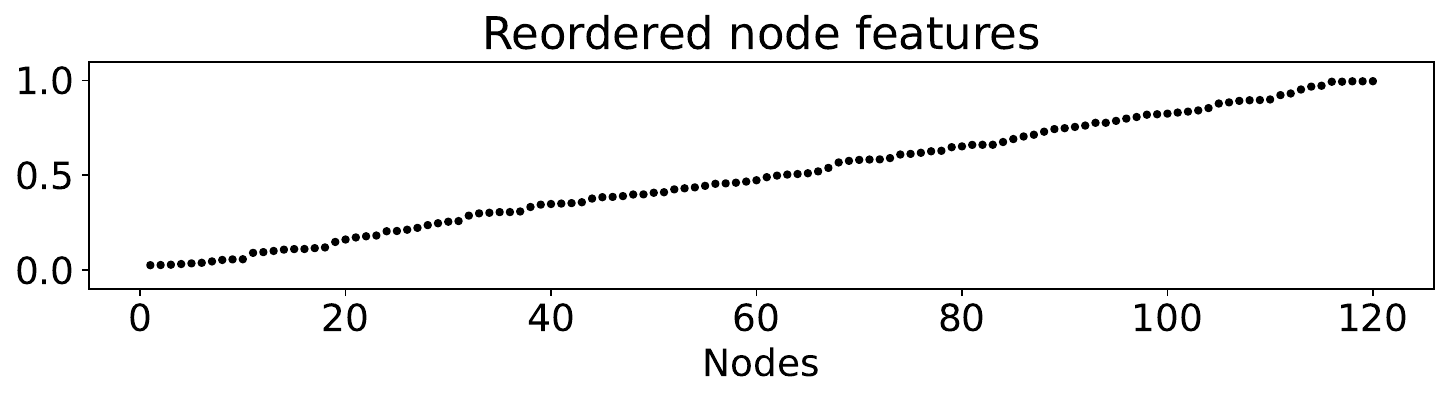}}
\caption{Matrices $\bar{A}$, $A$, $\underline{A}$, and $\underline{\hat{A}}$ and vectors $\bm{z}$ and $\underline{\bm{z}}$ (\textbf{undirected DGM}).}
\label{fig:syn2u}
\end{figure}

\begin{figure}[t]
\centerline{\includegraphics[width=0.24\hsize]{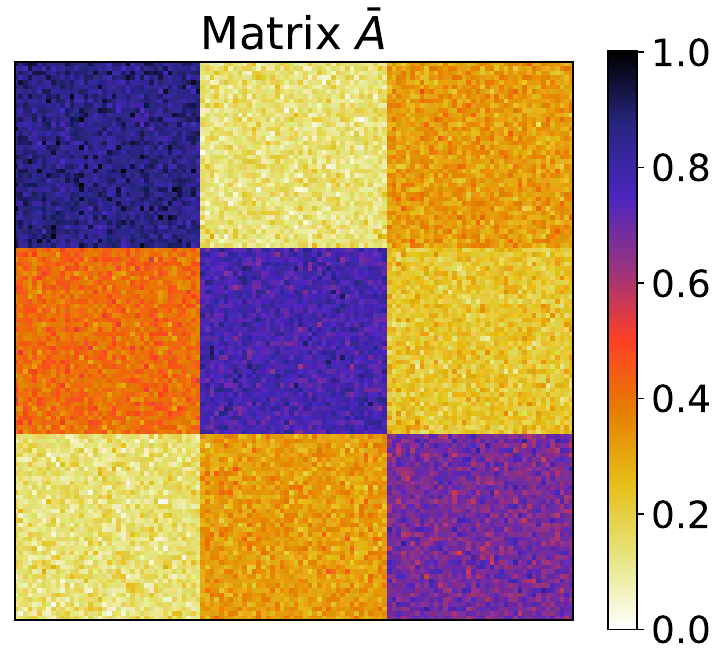}
\includegraphics[width=0.24\hsize]{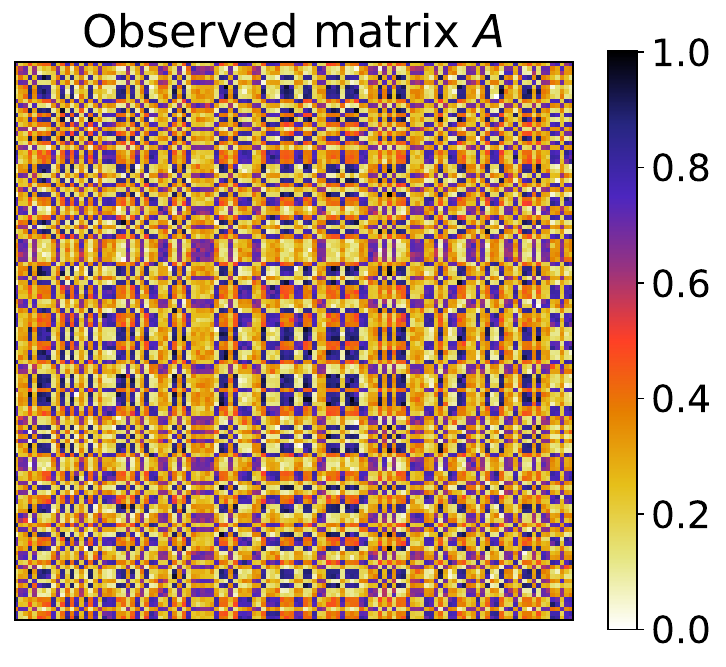}
\includegraphics[width=0.24\hsize]{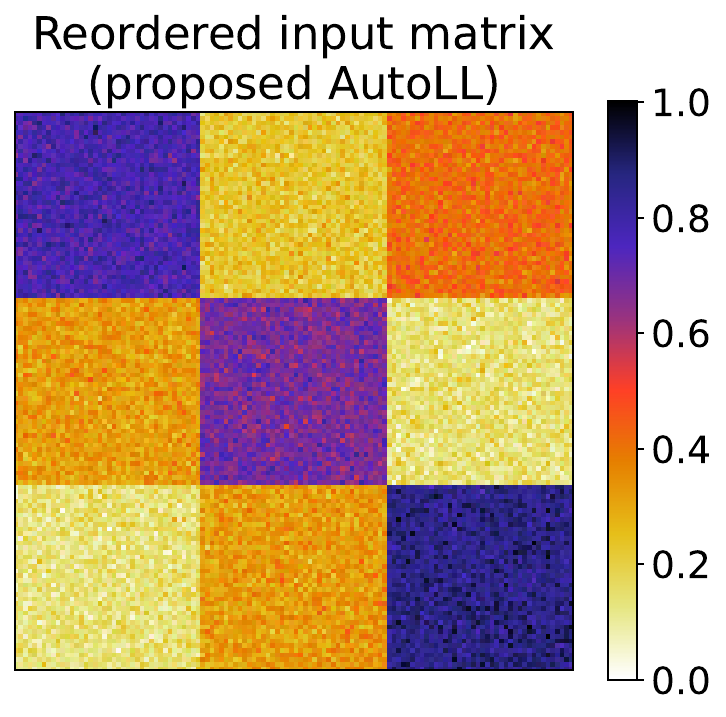}
\includegraphics[width=0.24\hsize]{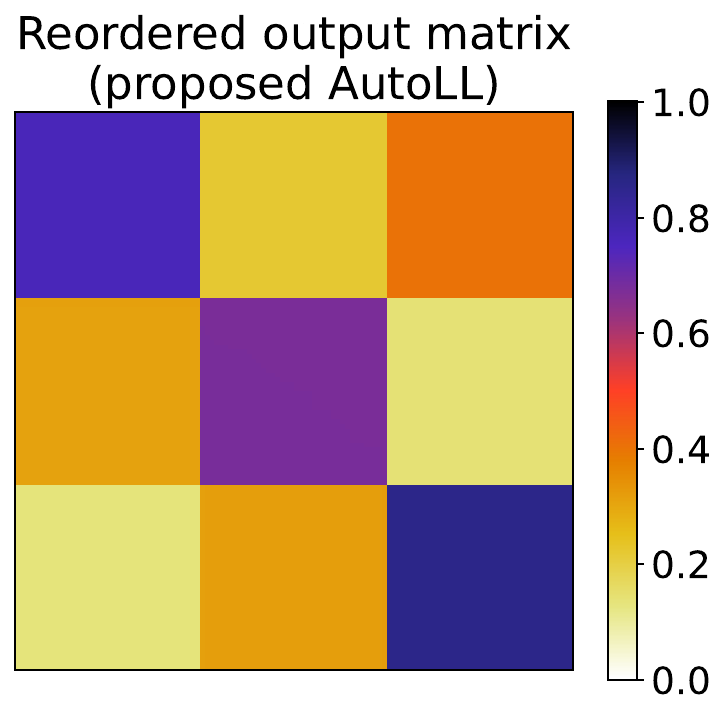}}
\centerline{\includegraphics[width=0.48\hsize]{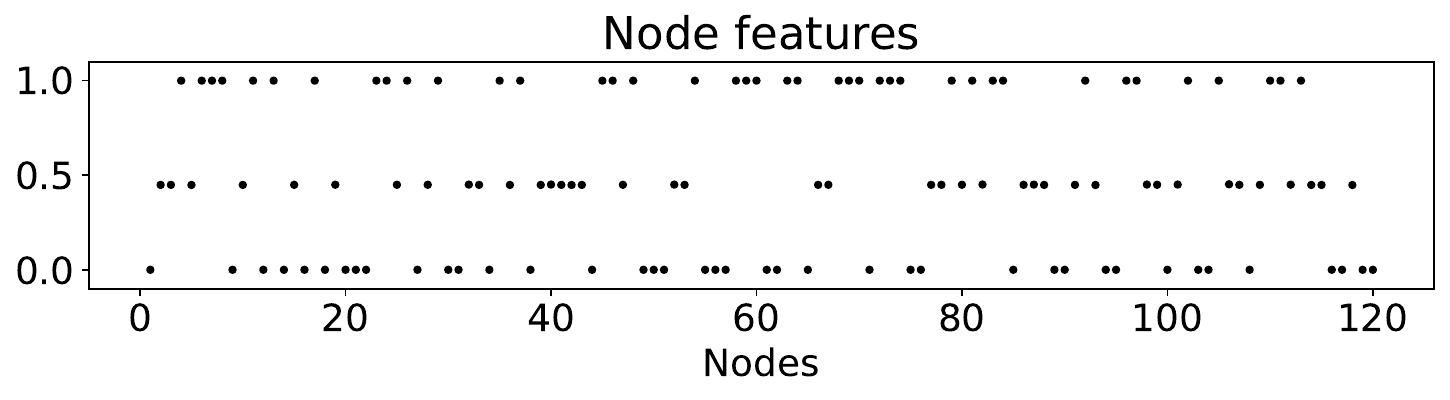}
\includegraphics[width=0.48\hsize]{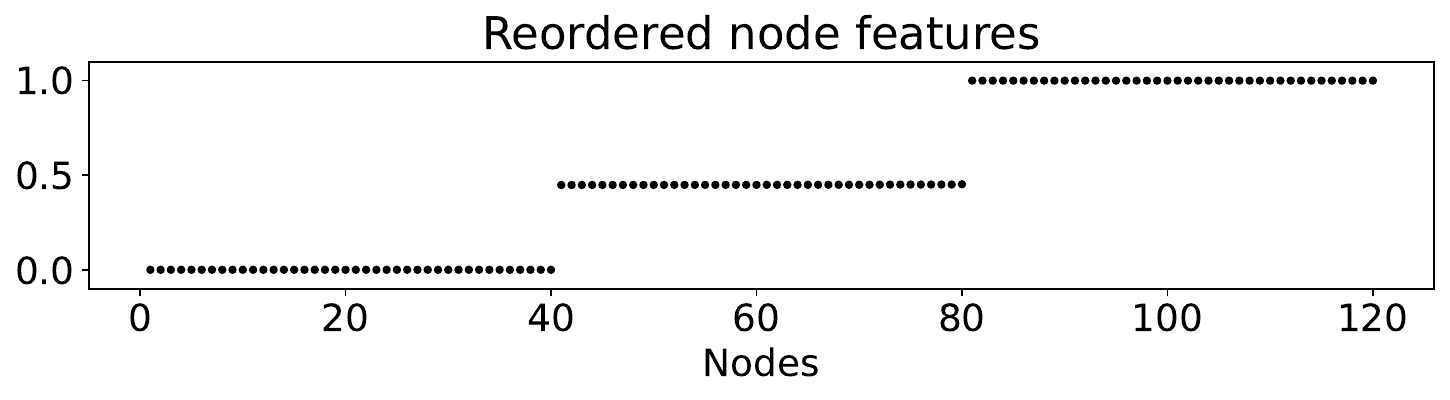}}
\caption{Matrices $\bar{A}$, $A$, $\underline{A}$, and $\underline{\hat{A}}$ and vectors $\bm{z}$ and $\underline{\bm{z}}$ (\textbf{directed SBM}).}\vspace{3mm}
\label{fig:syn1d}
\centerline{\includegraphics[width=0.24\hsize]{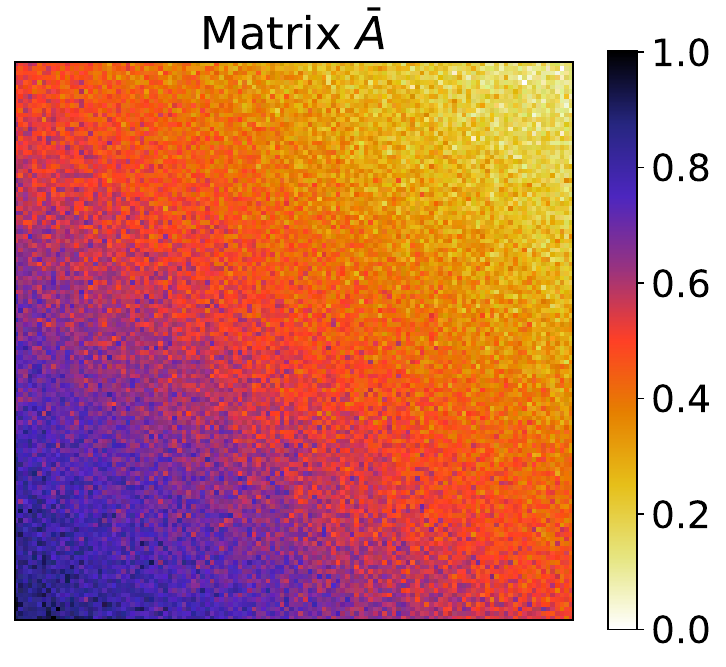}
\includegraphics[width=0.24\hsize]{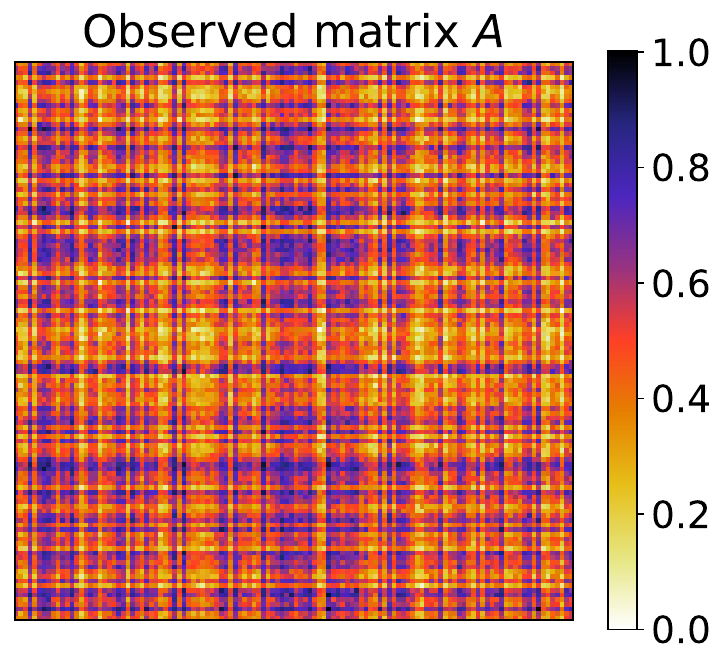}
\includegraphics[width=0.24\hsize]{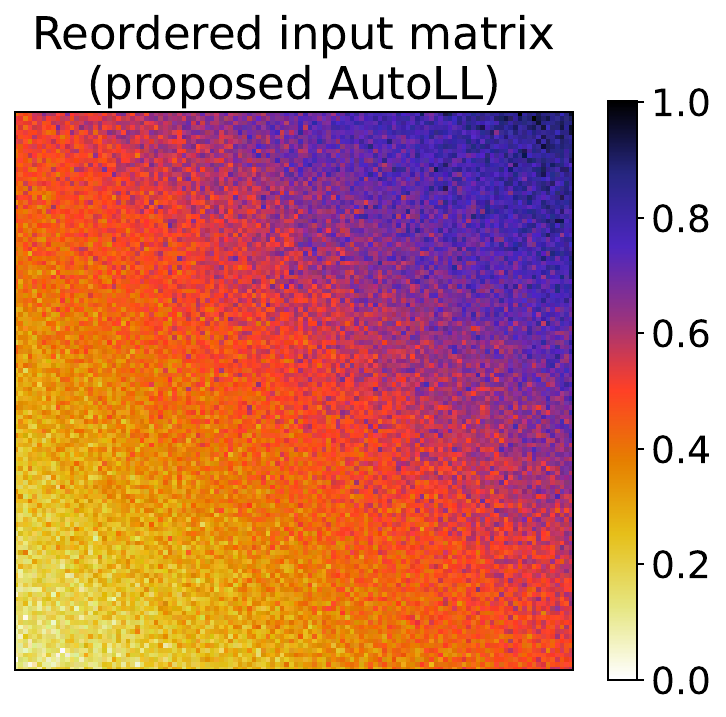}
\includegraphics[width=0.24\hsize]{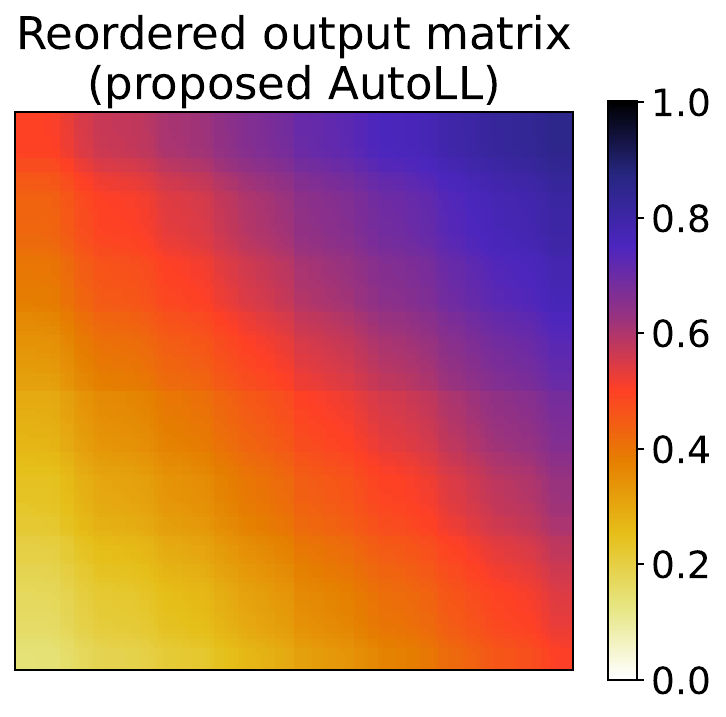}}
\centerline{\includegraphics[width=0.48\hsize]{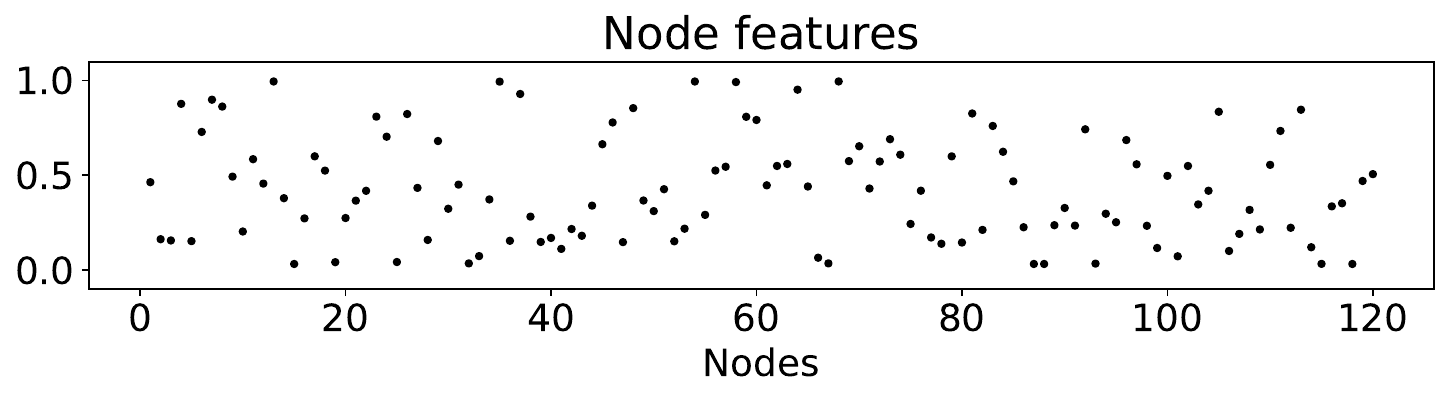}
\includegraphics[width=0.48\hsize]{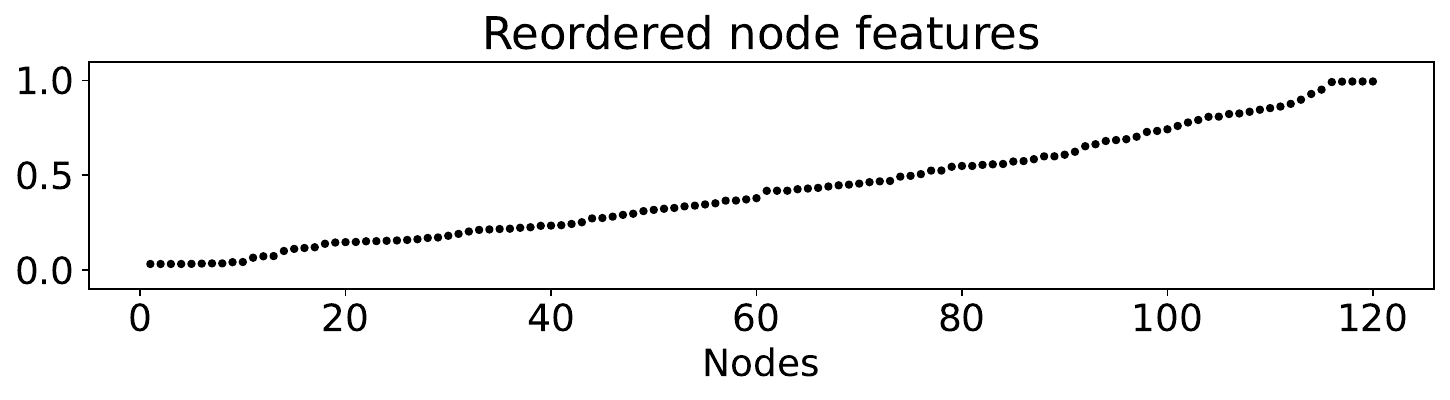}}
\caption{Matrices $\bar{A}$, $A$, $\underline{A}$, and $\underline{\hat{A}}$ and vectors $\bm{z}$ and $\underline{\bm{z}}$ (\textbf{directed DGM}).}
\label{fig:syn2d}
\end{figure}

Figs.~\ref{fig:syn1u}, \ref{fig:syn2u}, \ref{fig:syn1d}, and \ref{fig:syn2d} show the results of the undirected SBM, undirected DGM, directed SBM, and directed DGM, respectively. From the reordered observed and reconstructed matrices, $\underline{A}$ and $\underline{\hat{A}}$, respectively, we see that the proposed AutoLL models were able to successfully recover the underlying structure of a given adjacency matrix in all the settings, and that reconstructed matrix $\underline{\hat{A}}$ shows a denoised structural pattern in the observed matrix. It must be noted that in each model, there exist multiple ``correct'' orders of node indices. For instance, in SBM, the order of $K$ clusters and the node order within the same cluster are arbitrary. In DGM, the flipped order of the node indices in matrix $\bar{A}$ is also a correct order, as well as the original node order in matrix $\bar{A}$. 


\subsection{Objective evaluation of accuracy in graph reordering}
\label{sec:exp_compare}

Next, we compare the accuracy of the proposed AutoLL models to the existing spectral/dimension-reduction matrix reordering methods based on singular value decomposition (SVD) \cite{Friendly2002, Liu2003} and multidimensional scaling (MDS) \cite{Rodgers1992, Spence1974}. We use the same abbreviations as in the previous study \cite{Watanabe2021} for the three types of existing methods: SVD-Rank-One \cite{Liu2003}, SVD-Angle \cite{Friendly2002}, and MDS \cite{Rodgers1992, Spence1974}. 

\paragraph{SVD-Rank-One \cite{Liu2003}} In SVD-based methods, we assume that the observed matrix $A$ is generated based on the following bilinear model. $A = \bm{r} \bm{c}^{\top} + E$, where $\bm{r}$ and $\bm{c}$ are $n$-dimensional vectors and $E$ is an $n \times n$ residual matrix. An optimal solution $(\hat{\bm{r}}, \hat{\bm{c}})$ with the minimum reconstruction error, which is given by $\argmin_{(\bm{r}, \bm{c}) \in \mathbb{R}^n \times \mathbb{R}^n} \| A - \bm{r} \bm{c}^{\top} \|_{\mathrm{F}}^2$, is $\hat{\bm{r}} = \sqrt{\lambda_1} \bm{u}_1$ and $\hat{\bm{c}} = \sqrt{\lambda_1} \bm{v}_1$, where $\lambda_1$, $\bm{u}_1 \in \mathbb{R}^n$, and $\bm{v}_1 \in \mathbb{R}^n$ are the largest singular value of matrix $A$ and the corresponding row and column singular vectors, respectively. In the SVD-Rank-One process, we adopt the ascending order of the entries in vector $\hat{\bm{r}}$ as the estimated order of node indices for graph reordering. 

\paragraph{SVD-Angle \cite{Friendly2002}} In this method, we define the node order based on the information of the top two singular vectors. Given the observed adjacency matrix $A$, we apply SVD to the row-wise normalized matrix $\tilde{A}$, which is given by
\begin{align}
\tilde{A} = (\tilde{A}_{ij})_{1 \leq i, j \leq n}, \ \ \ 
\tilde{A}_{ij} = \frac{\tilde{A}^{(0)}_{ij}}{\sqrt{\frac{1}{n} \sum_{j = 1}^n \left( \tilde{A}^{(0)}_{ij} \right)^2}},  
\end{align}
where $\tilde{A}^{(0)}_{ij} = A_{ij} - \frac{1}{n} \sum_{j = 1}^n A_{ij}$. We define the angle $\alpha_i$ between the top two row singular vectors, $\bm{u}_1$ and $\bm{u}_2$, as follows. 
\begin{align}
\alpha_i = \tan^{-1} (u_{i2} / u_{i1}) + \pi I[u_{i1} \leq 0], 
\end{align}
where $I[\cdot]$ is an indicator function. To convert the set of angles $\{\alpha_i\}$ to the node order, we split it at the largest gap between two adjacent angles. Specifically, we first define the ascending order $\rho^{(0)}$ of the angles $\alpha_{\rho^{(0)} (1)} \leq \dots \leq \alpha_{\rho^{(0)} (n)}$. Then, we compute the gaps between two adjacent angles as $d_1 = 2\pi + \alpha_{\rho^{(0)} (1)} - \alpha_{\rho^{(0)} (n)}$ and $d_i = \alpha_{\rho^{(0)} (i)} - \alpha_{\rho^{(0)} (i-1)}$ for $i = 2, \dots, n$. Let $i' = \argmax_{i \in \{1, \dots, n\}} d_i$ (i.e., the node index with the largest gap in angle with an adjacency node). By splitting the node indices at the $i'$th node, we obtain the final node order $\rho$. For all $i \in \{1, \dots, n\}$, 
\begin{align}
\rho (i) = \begin{cases}
\rho^{(0)} (i) + n - i' + 1 & \mathrm{if}\ \rho^{(0)} (i) < i',\\
\rho^{(0)} (i) - i' + 1 & \mathrm{otherwise}. 
\end{cases}
\end{align}

\paragraph{MDS \cite{Rodgers1992, Spence1974}} In MDS, we do not apply SVD directly to the original adjacency matrix $A$ but to the following semi-positive definite matrix $\tilde{B}$. 
\begin{align}
\label{eq:def_B}
\tilde{B} = -\frac{1}{2} \left( I - n^{-1} Q \right) \tilde{D} \left( I - n^{-1} Q \right), 
\end{align}
where
\begin{align}
&\tilde{D} = (\tilde{D}_{ii'})_{1 \leq i, i' \leq n}, \nonumber \\
&\tilde{D}_{ii'} = \sum_{j = 1}^n (A_{ij} - A_{i'j})^2, \ \ \ i, i' = 1, \dots, n, \nonumber \\
&Q = (Q_{ii'})_{1 \leq i, i' \leq n}, \ \ \ Q_{ii'} = 1, \ \ \ i, i' = 1, \dots, n. 
\end{align}
Matrix $\tilde{B}$ in \eqref{eq:def_B} reflects the proximity between pairs of rows in the adjacency matrix $A$. Let $\tilde{B} = UDU^{\top}$ be the singular value decomposition of matrix $\tilde{B}$, where $U = (U_{ij})_{1 \leq i, j \leq n} \in \mathbb{R}^{n \times n}$ is an orthogonal matrix, and $D$ is a diagonal matrix with singular values. We use the ascending order of the entries in vector $\bm{u}_1 = \begin{bmatrix} U_{11} & \cdots & U_{n1} \end{bmatrix}^{\top}$ (i.e., the eigenvector of matrix $\tilde{B}$ corresponding to the maximum eigenvalue) as the order of node indices. 

In both directed and undirected cases, we generated adjacency matrices based on the DGM with the same mean matrix $B$ as in Section \ref{sec:exp_syn} and the following $10$ patterns of standard deviations: $\sigma_t = 0.03t$ for $t = 1, \dots, 10$. For each $t$, we defined $10$ adjacency matrices $A$ based on the same procedure as in Section \ref{sec:exp_syn} and applied the proposed and existing methods to them. Because the training result of the AutoLL model depends on the random parameter initialization and mini-batch selection, we trained an AutoLL model with the same matrix $A$ $10$ times and selected the model with the minimum average training loss for the last $100$ iterations. 

To evaluate the accuracy of each graph reordering method, we adopted the criterion of matrix reordering error, which was introduced for two-mode matrix reordering in a previous study \cite{Watanabe2021}. In our case, the graph reordering error can be defined as the mean squared error between the entries of the mean matrices $B$ with the correct and estimated node orders. Because the flipped node order of the correct ordering is also correct, as explained in Section \ref{sec:exp_syn}, we chose the solution with the minimum error from the two (i.e., original and flipped) estimated node orders and computed the final graph reordering error based on it for each method. 

\begin{figure}[p]
\centerline{\includegraphics[width=0.9\hsize]{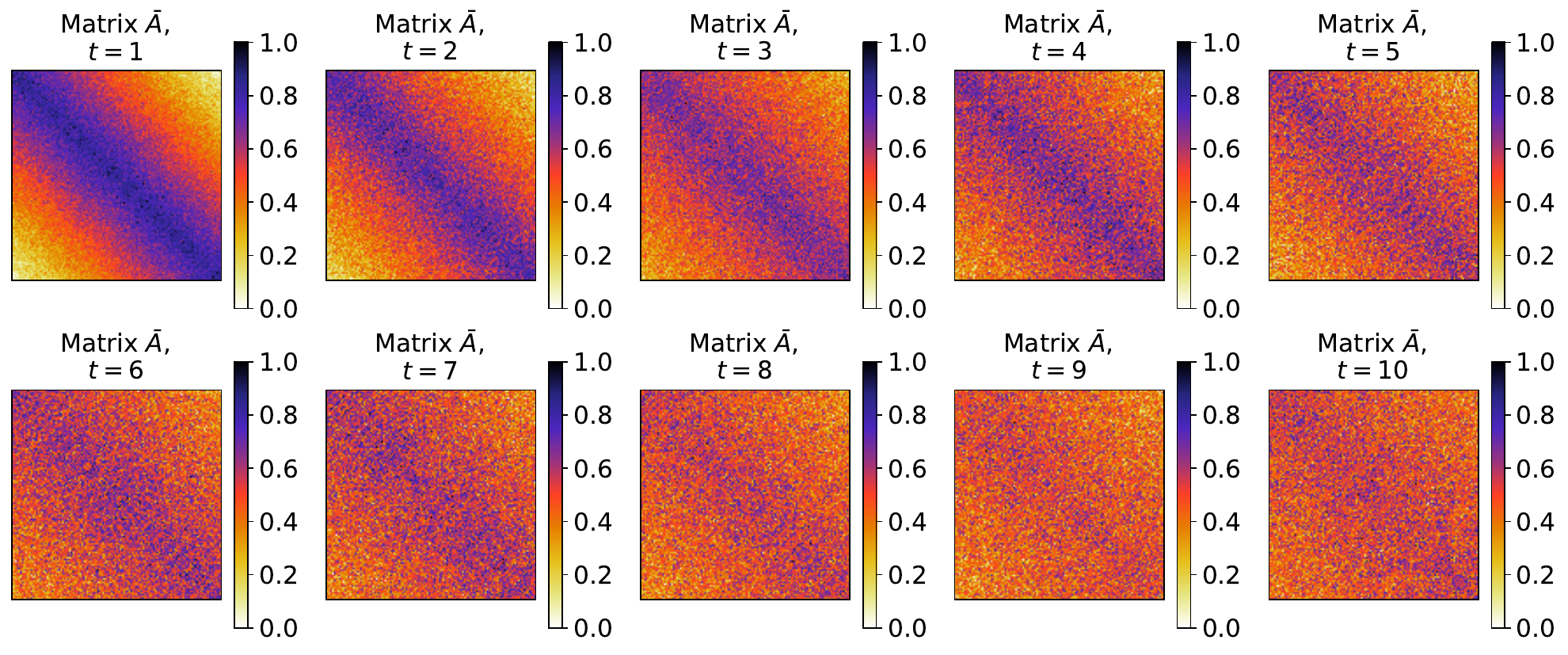}}
\caption{Examples of matrices $\bar{A}$ for different settings of $t$ (\textbf{undirected case}).}\vspace{3mm}
\label{fig:compare_u_bar_A}
\centerline{\includegraphics[width=0.9\hsize]{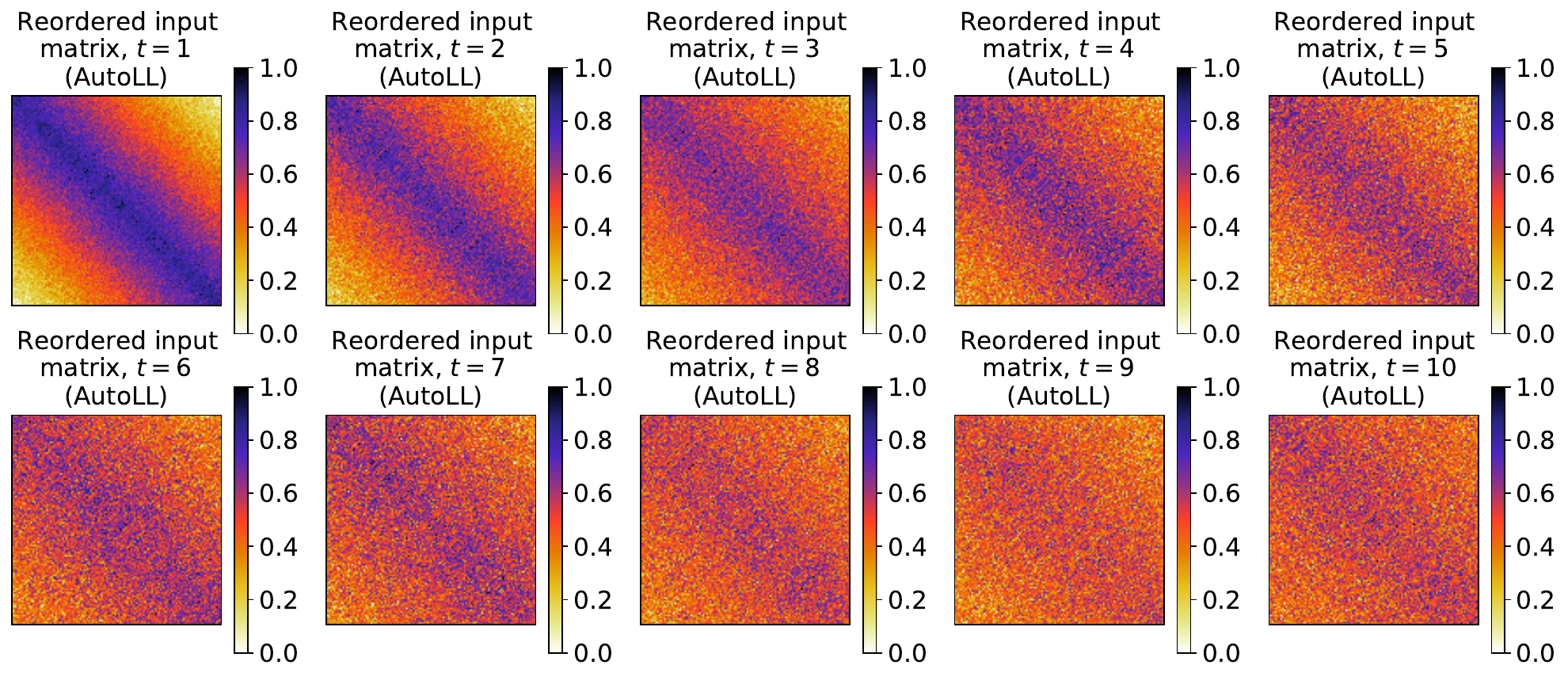}}
\caption{Examples of matrices $\underline{A}$ of the proposed AutoLL for different settings of $t$ (\textbf{undirected case}).}\vspace{3mm}
\label{fig:compare_u_underline_A_AutoLL}
\centerline{\includegraphics[width=0.9\hsize]{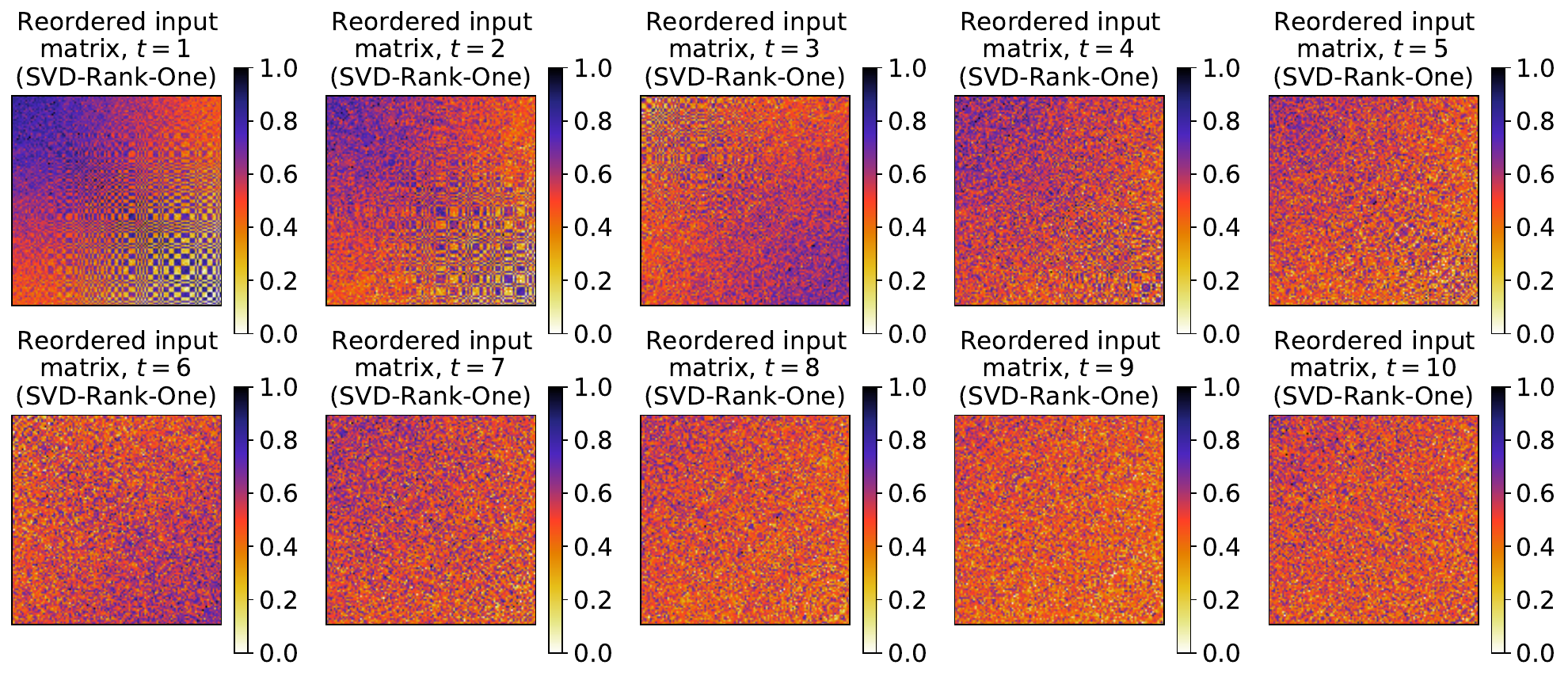}}
\caption{Examples of matrices $\underline{A}$ of SVD-Rank-One for different settings of $t$ (\textbf{undirected case}).}\vspace{3mm}
\label{fig:compare_u_underline_A_SVD_Rank_One}
\end{figure}

\begin{figure}[t]
\centerline{\includegraphics[width=0.9\hsize]{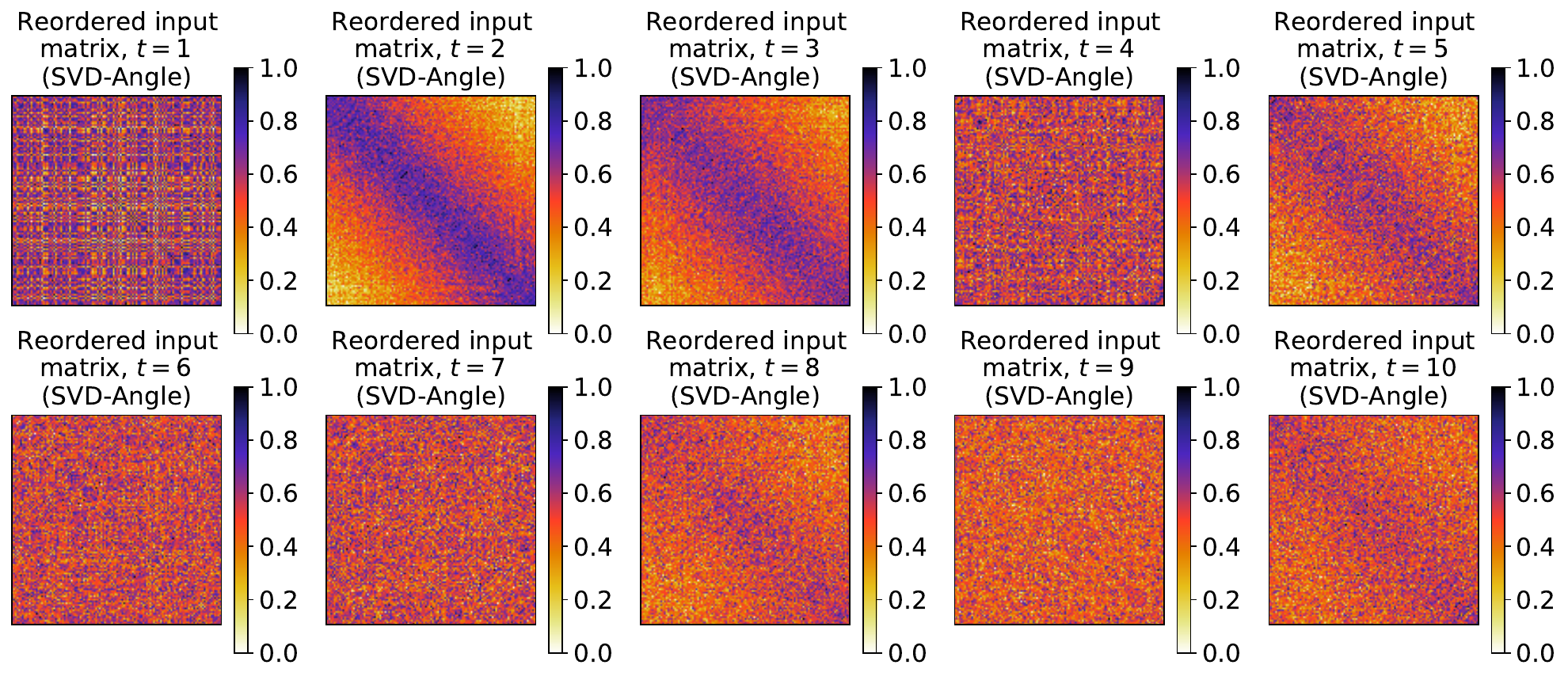}}
\caption{Examples of matrices $\underline{A}$ of SVD-Angle for different settings of $t$ (\textbf{undirected case}).}\vspace{3mm}
\label{fig:compare_u_underline_A_SVD_Angle}
\centerline{\includegraphics[width=0.9\hsize]{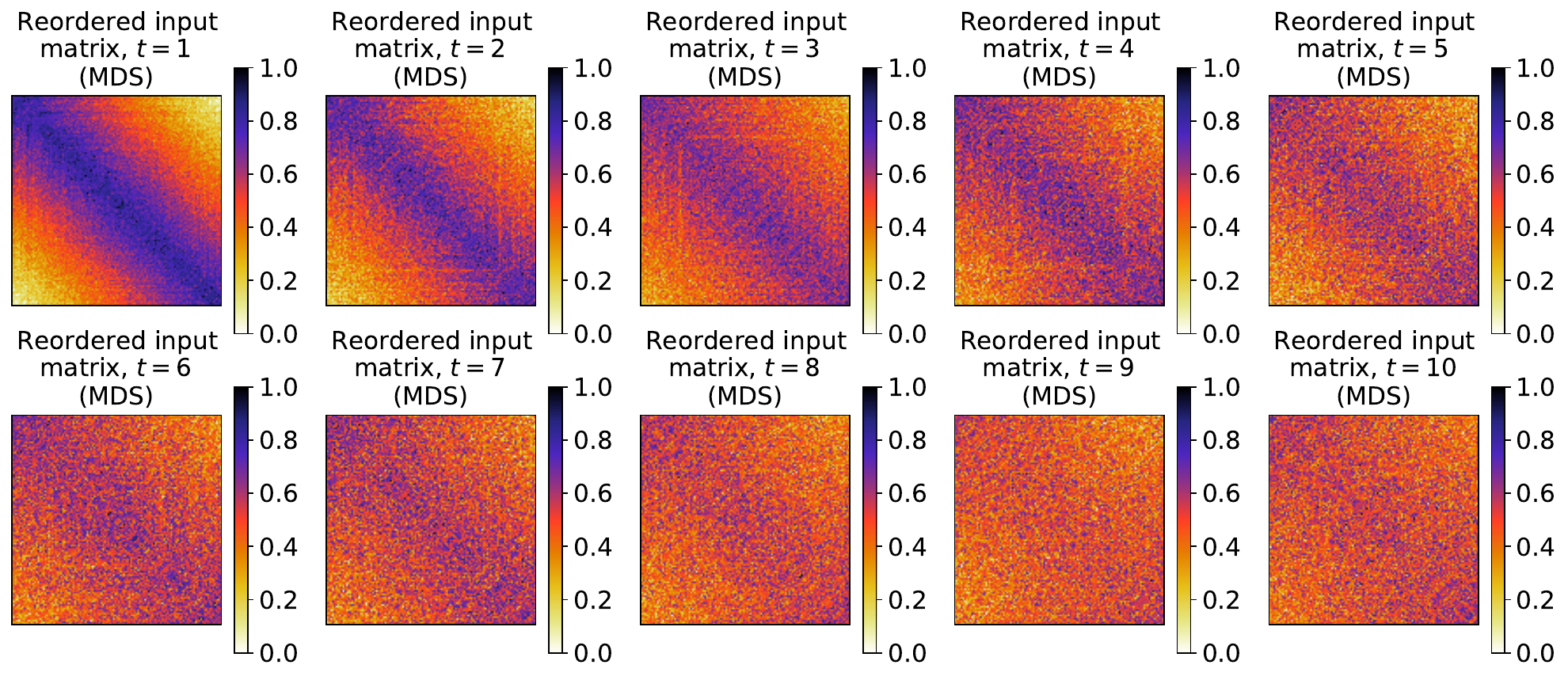}}
\caption{Examples of matrices $\underline{A}$ of MDS for different settings of $t$ (\textbf{undirected case}).}\vspace{3mm}
\label{fig:compare_u_underline_A_MDS}
\end{figure}

\begin{figure}[p]
\centerline{\includegraphics[width=0.9\hsize]{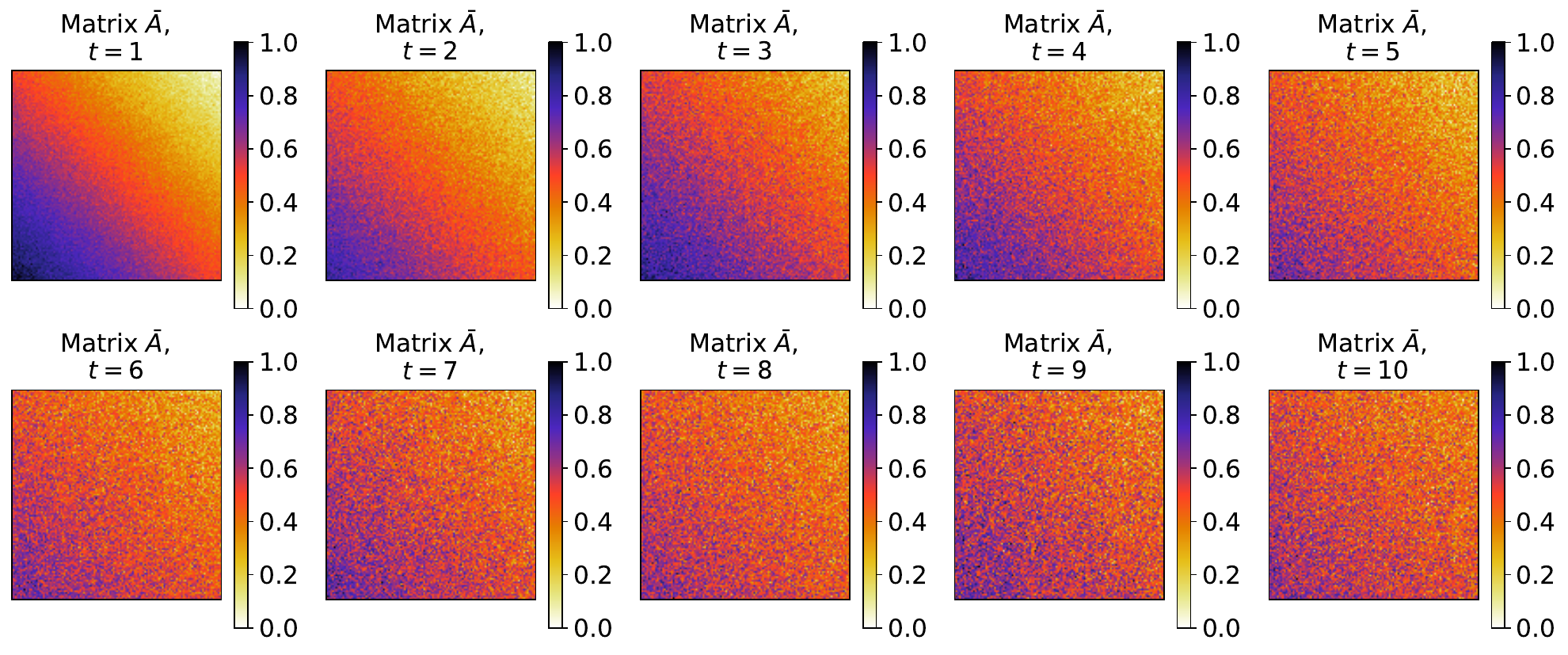}}
\caption{Examples of matrices $\bar{A}$ for different settings of $t$ (\textbf{directed case}).}\vspace{3mm}
\label{fig:compare_d_bar_A}
\centerline{\includegraphics[width=0.9\hsize]{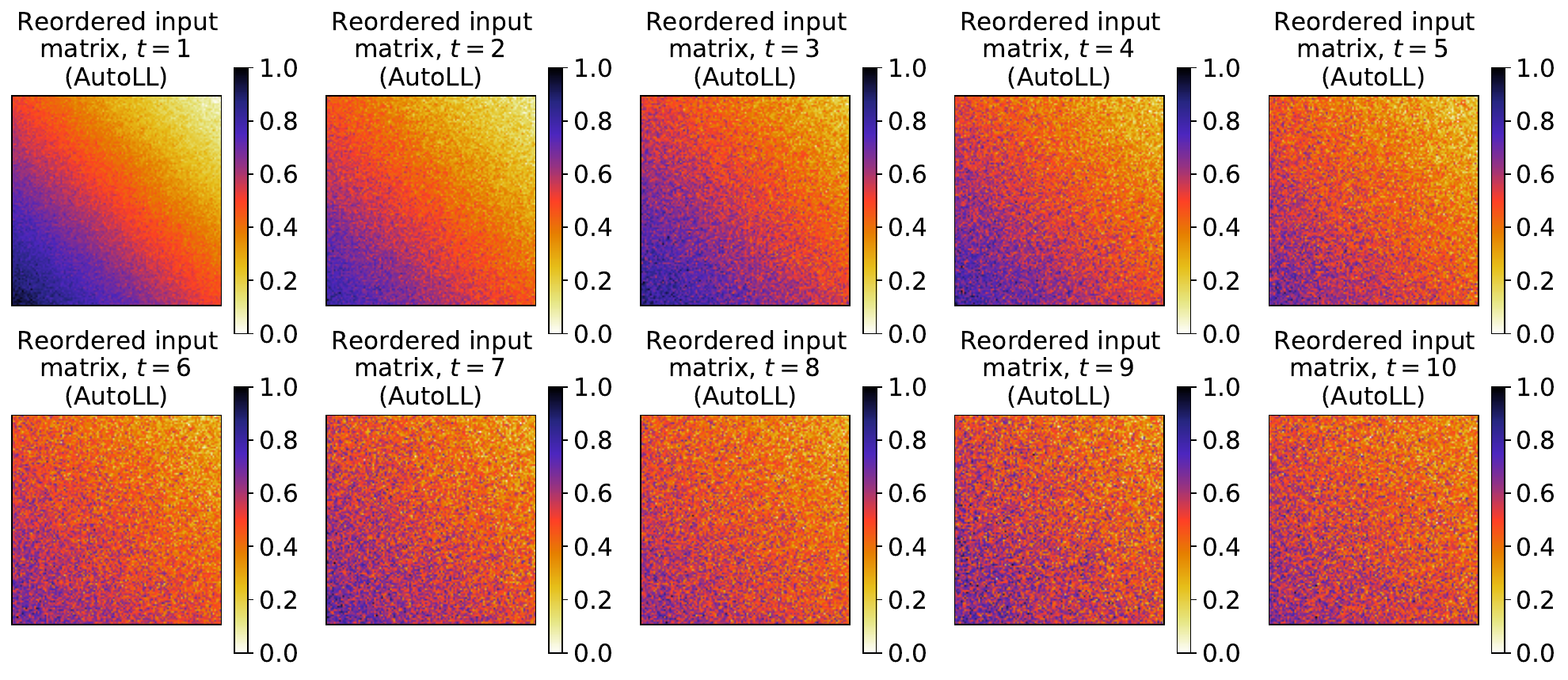}}
\caption{Examples of matrices $\underline{A}$ of the proposed AutoLL for different settings of $t$ (\textbf{directed case}).}\vspace{3mm}
\label{fig:compare_d_underline_A_AutoLL}
\centerline{\includegraphics[width=0.9\hsize]{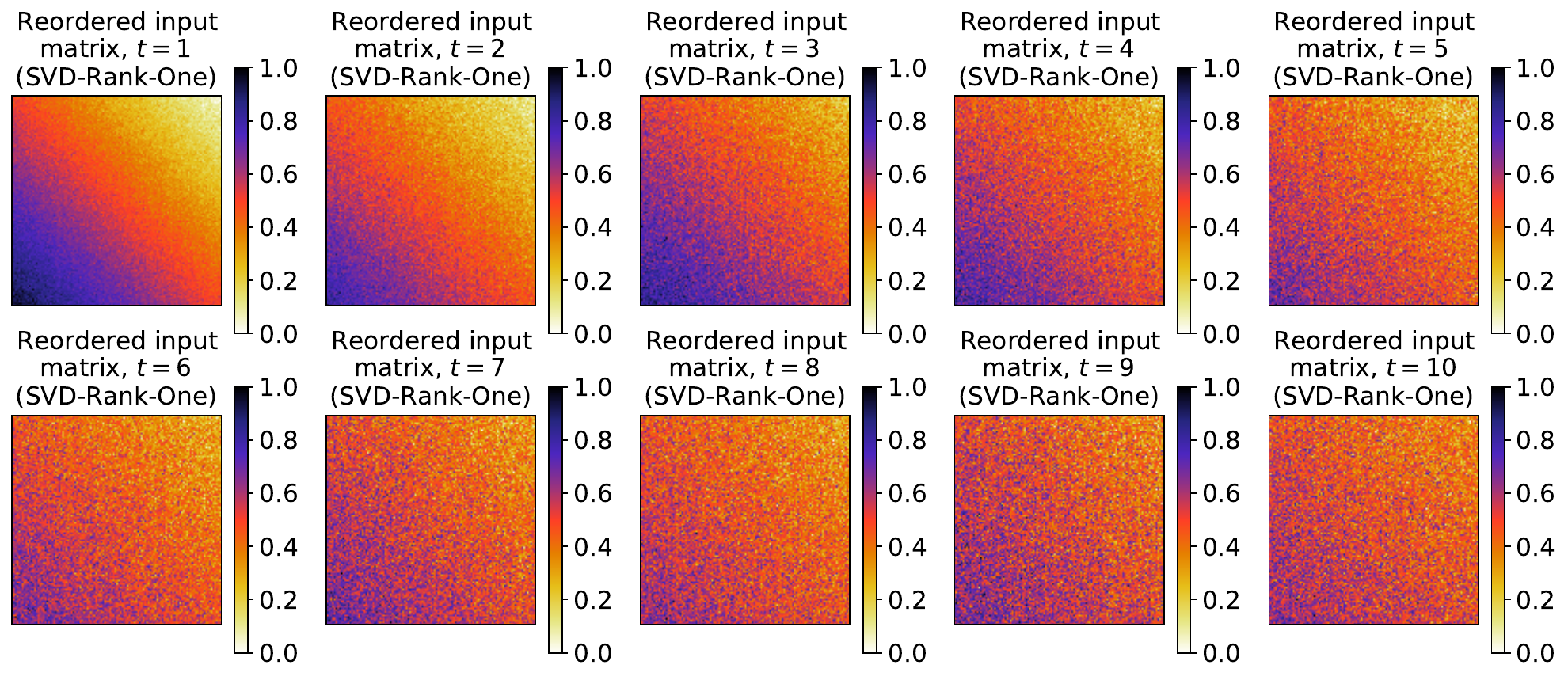}}
\caption{Examples of matrices $\underline{A}$ of SVD-Rank-One for different settings of $t$ (\textbf{directed case}).}\vspace{3mm}
\label{fig:compare_d_underline_A_SVD_Rank_One}
\end{figure}

\begin{figure}[t]
\centerline{\includegraphics[width=0.9\hsize]{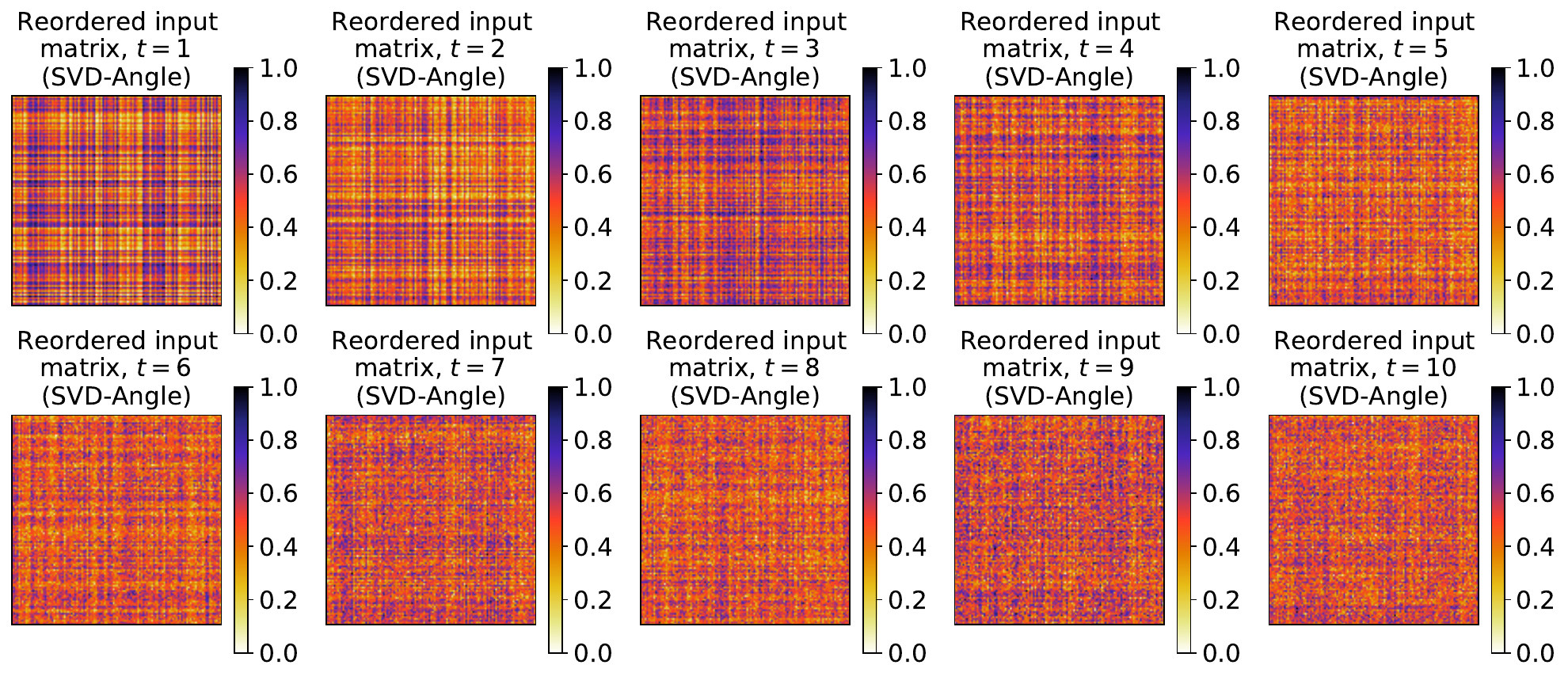}}
\caption{Examples of matrices $\underline{A}$ of SVD-Angle for different settings of $t$ (\textbf{directed case}).}\vspace{3mm}
\label{fig:compare_d_underline_A_SVD_Angle}
\centerline{\includegraphics[width=0.9\hsize]{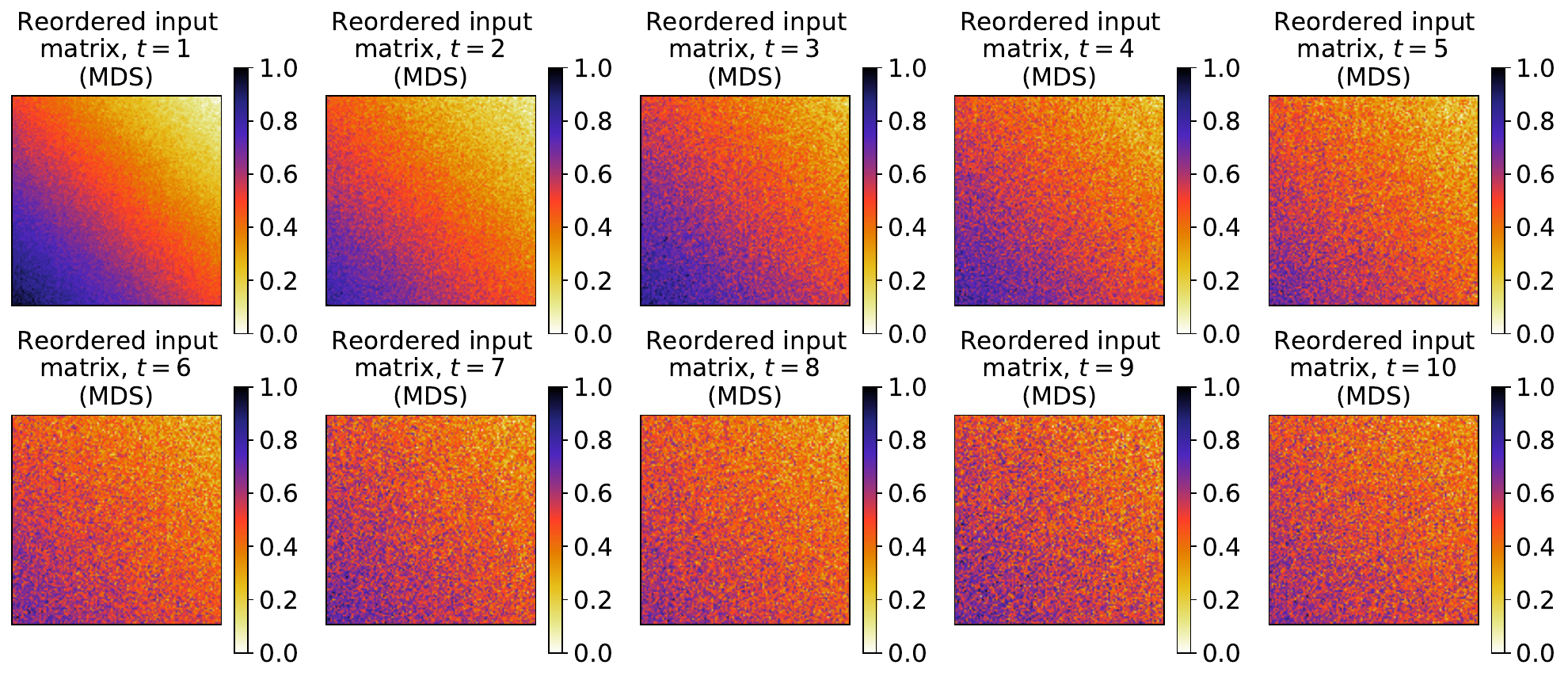}}
\caption{Examples of matrices $\underline{A}$ of MDS for different settings of $t$ (\textbf{directed case}).}\vspace{3mm}
\label{fig:compare_d_underline_A_MDS}
\end{figure}

\begin{figure}[t]
\centerline{\includegraphics[width=0.6\hsize]{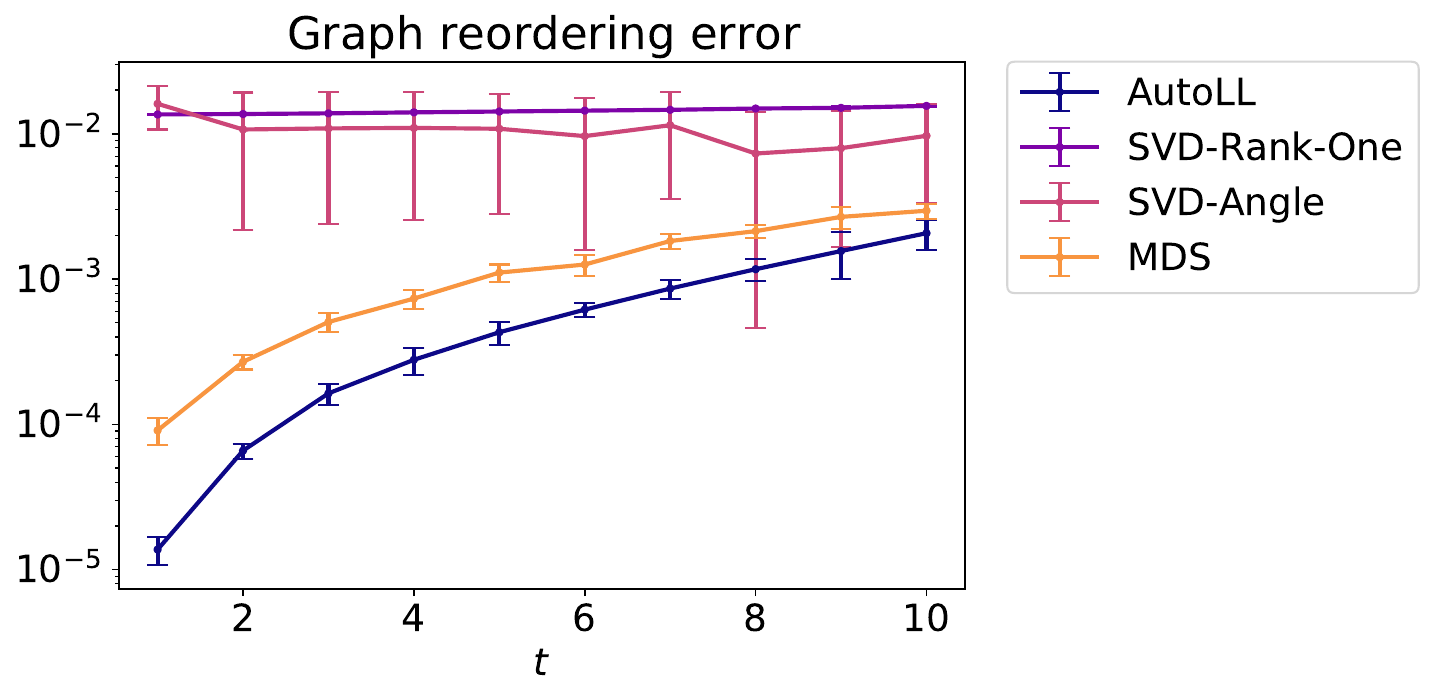}}
\caption{Average graph reordering errors for $10$ trials (\textbf{undirected case}). Error bars indicate sample standard deviation.}\vspace{3mm}
\label{fig:compare_u}
\centerline{\includegraphics[width=0.6\hsize]{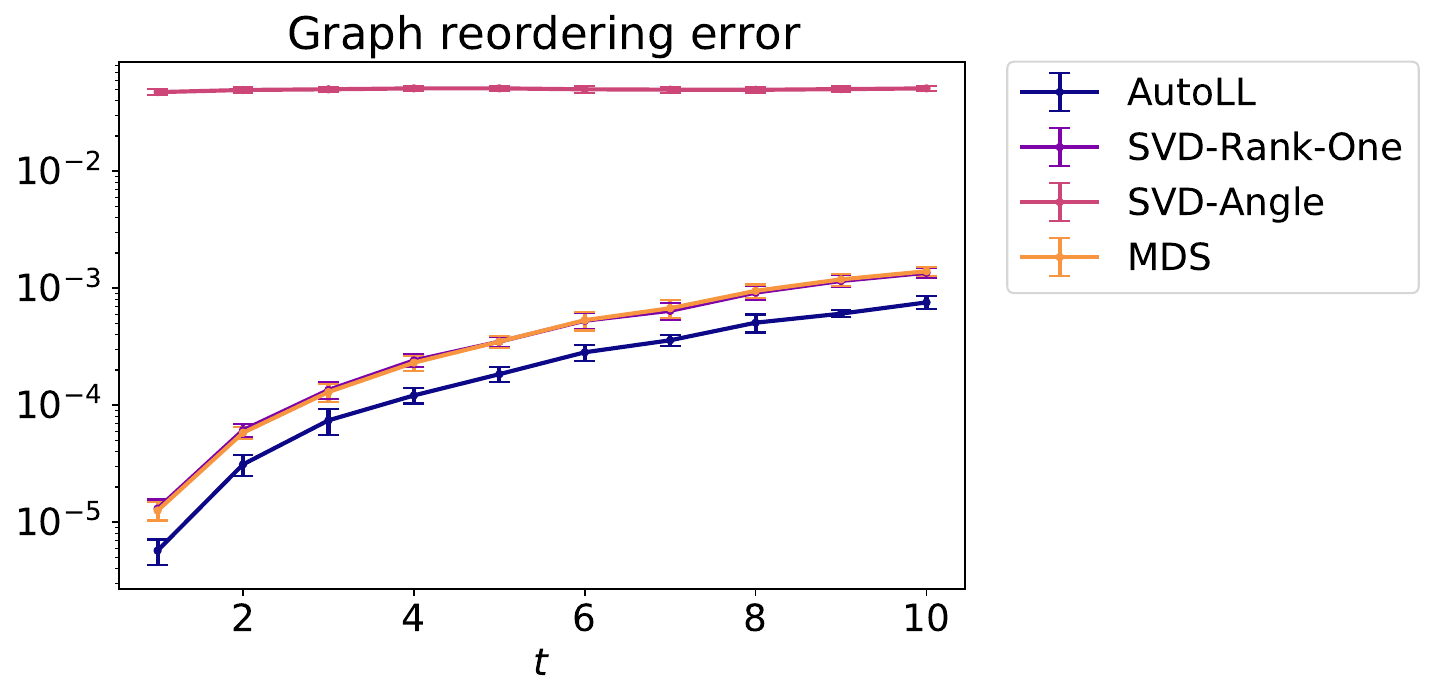}}
\caption{Average graph reordering errors for $10$ trials (\textbf{directed case}). Error bars indicate sample standard deviation.}
\label{fig:compare_d}
\end{figure}

Fig.~\ref{fig:compare_u_bar_A} (\ref{fig:compare_d_bar_A}) shows the examples of the original adjacency matrices $\bar{A}$ with the correct node orders, and Figs.~\ref{fig:compare_u_underline_A_AutoLL}, \ref{fig:compare_u_underline_A_SVD_Rank_One}, \ref{fig:compare_u_underline_A_SVD_Angle}, and \ref{fig:compare_u_underline_A_MDS} (\ref{fig:compare_d_underline_A_AutoLL}, \ref{fig:compare_d_underline_A_SVD_Rank_One}, \ref{fig:compare_d_underline_A_SVD_Angle}, and \ref{fig:compare_d_underline_A_MDS}), respectively, show those of the reordered observed matrices of the proposed AutoLL, SVD-Rank-One, SVD-Angle, and MDS in the undirected (directed) newtork settings. 
Figs.~\ref{fig:compare_u} and \ref{fig:compare_d}, respectively, show the graph reordering errors with different settings of $t$ in undirected and directed cases. From these figures, in both directed and undirected cases, the proposed AutoLL achieved the best average performance among the four methods. 


\subsection{Robustness to the outliers}
\label{sec:exp_outlier}

We also check the robustness of the proposed and conventional graph reordering methods for the outliers. As in Section \ref{sec:exp_compare}, we generated adjacency matrices based on the directed and undirected DGMs with the same mean matrix $B$ as in Section \ref{sec:exp_syn} and the standard deviation $\sigma = 0.03$. Then, we replaced each entry $\bar{A}^{(0)}_{ij}$ of the generated matrix $\bar{A}^{(0)}$ with zero with a probability of $0.01t$, where $t = 1, \dots, 10$. 
For each $t$, we defined $10$ adjacency matrices $A$ based on the same procedure as in Section \ref{sec:exp_syn} and applied AutoLL, SVD-Rank-One, SVD-Angle, and MDS to them. As in Section \ref{sec:exp_compare}, we trained an AutoLL model with the same matrix $A$ $10$ times and selected the model with the minimum average training loss for the last $100$ iterations. 

\begin{figure}[p]
\centerline{\includegraphics[width=0.9\hsize]{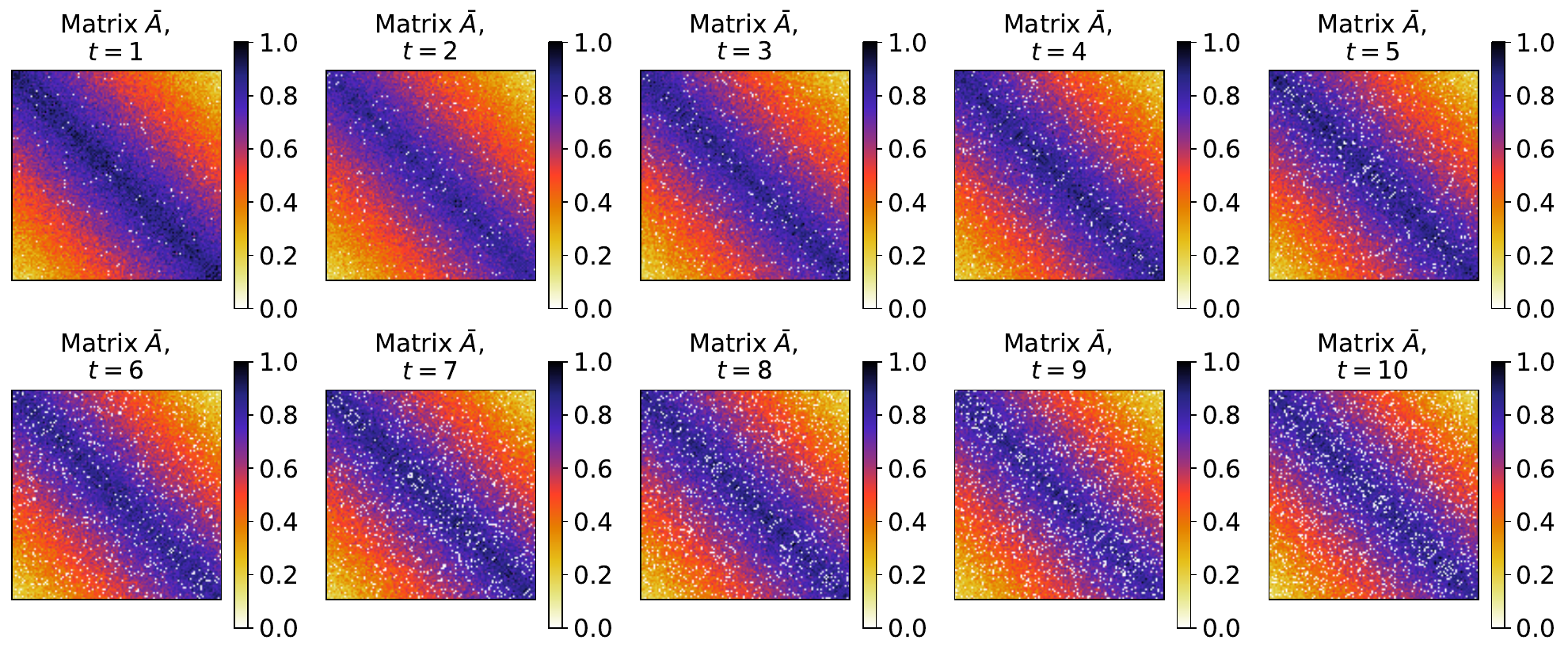}}
\caption{Examples of matrices $\bar{A}$ for different settings of $t$ (\textbf{undirected case with outliers}).}\vspace{3mm}
\label{fig:compare_u_bar_A_outlier}
\centerline{\includegraphics[width=0.9\hsize]{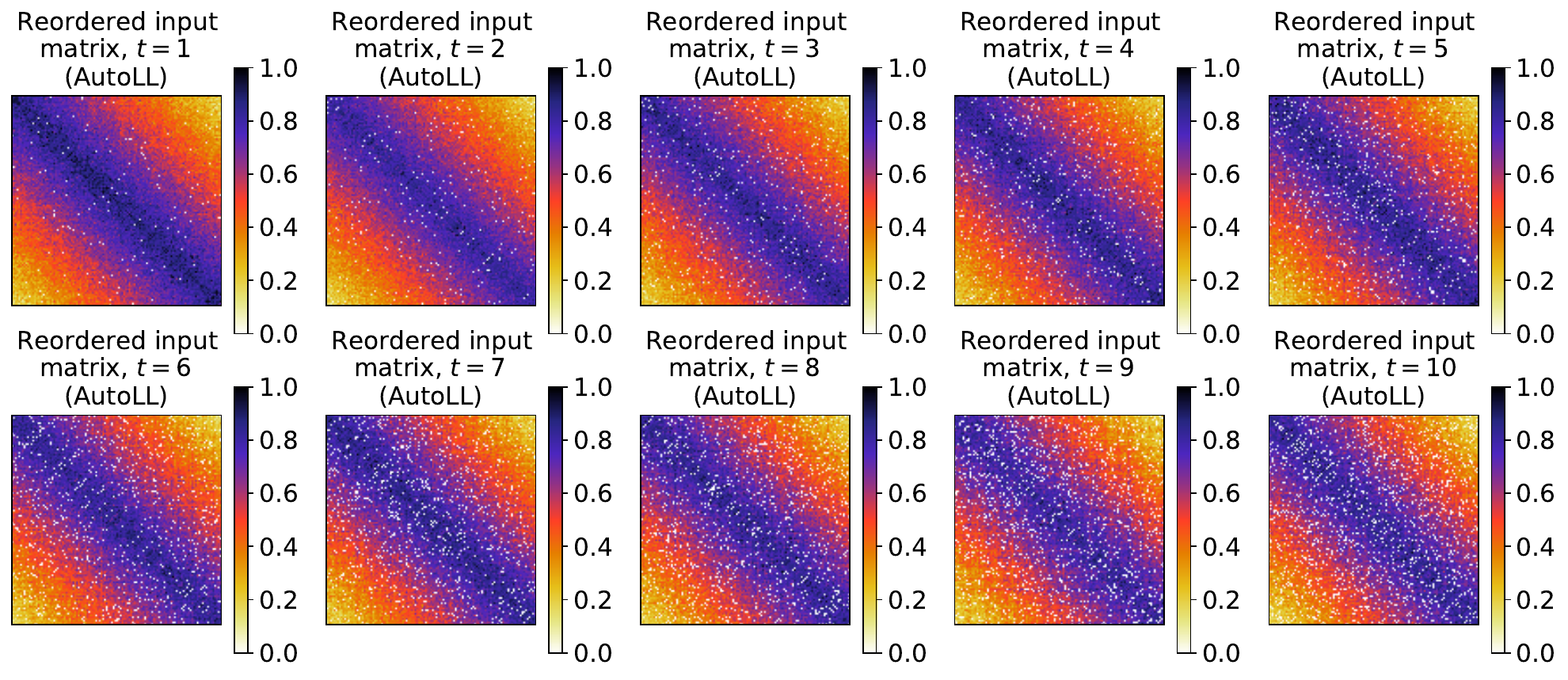}}
\caption{Examples of matrices $\underline{A}$ of the proposed AutoLL for different settings of $t$ (\textbf{undirected case with outliers}).}\vspace{3mm}
\label{fig:compare_u_underline_A_AutoLL_outlier}
\centerline{\includegraphics[width=0.9\hsize]{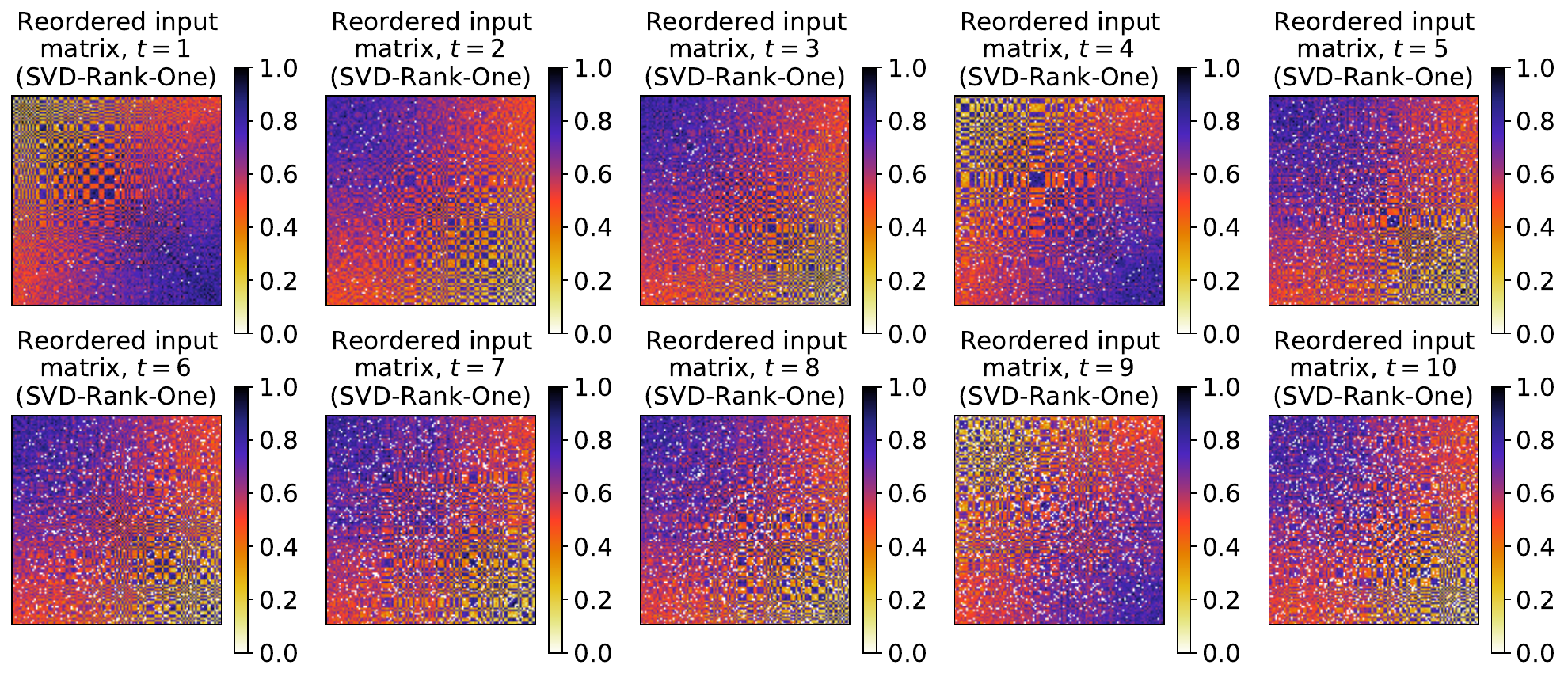}}
\caption{Examples of matrices $\underline{A}$ of SVD-Rank-One for different settings of $t$ (\textbf{undirected case with outliers}).}\vspace{3mm}
\label{fig:compare_u_underline_A_SVD_Rank_One_outlier}
\end{figure}

\begin{figure}[t]
\centerline{\includegraphics[width=0.9\hsize]{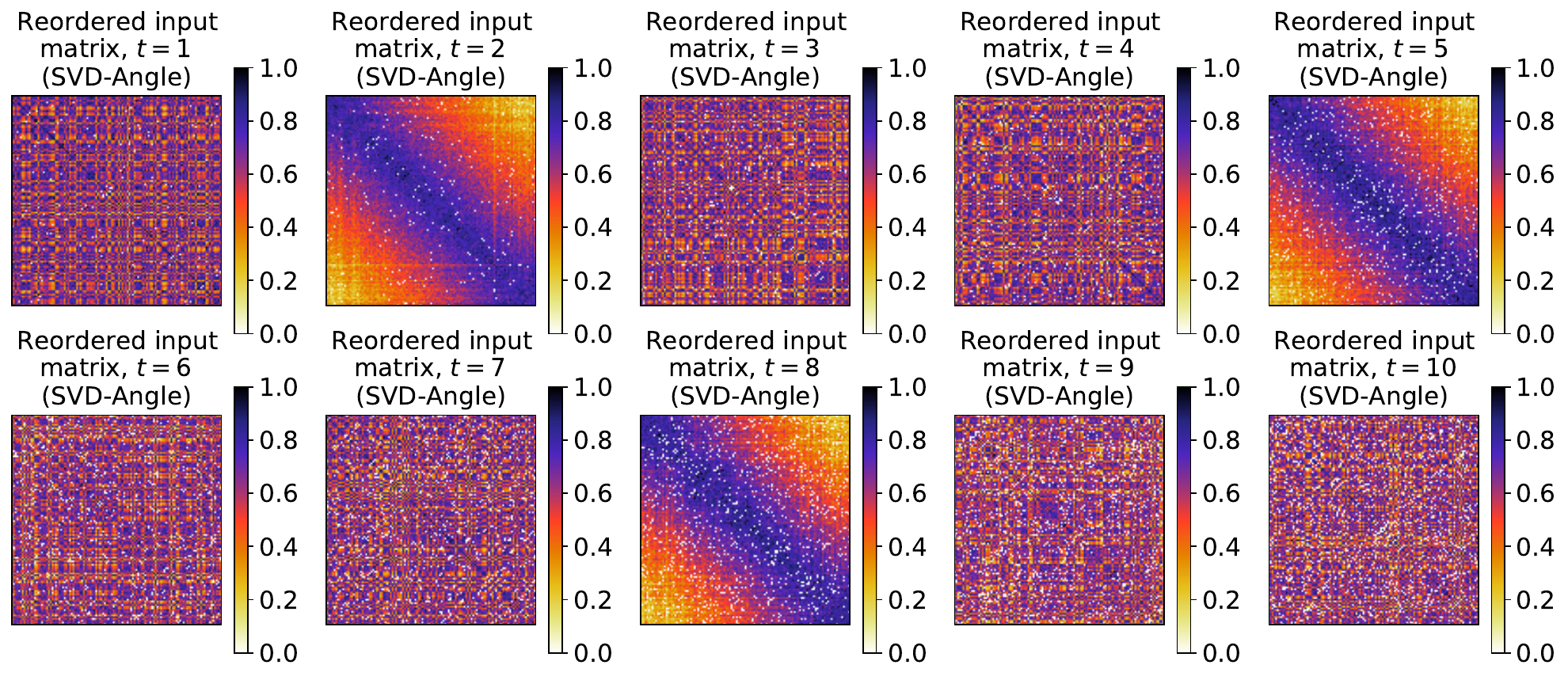}}
\caption{Examples of matrices $\underline{A}$ of SVD-Angle for different settings of $t$ (\textbf{undirected case with outliers}).}\vspace{3mm}
\label{fig:compare_u_underline_A_SVD_Angle_outlier}
\centerline{\includegraphics[width=0.9\hsize]{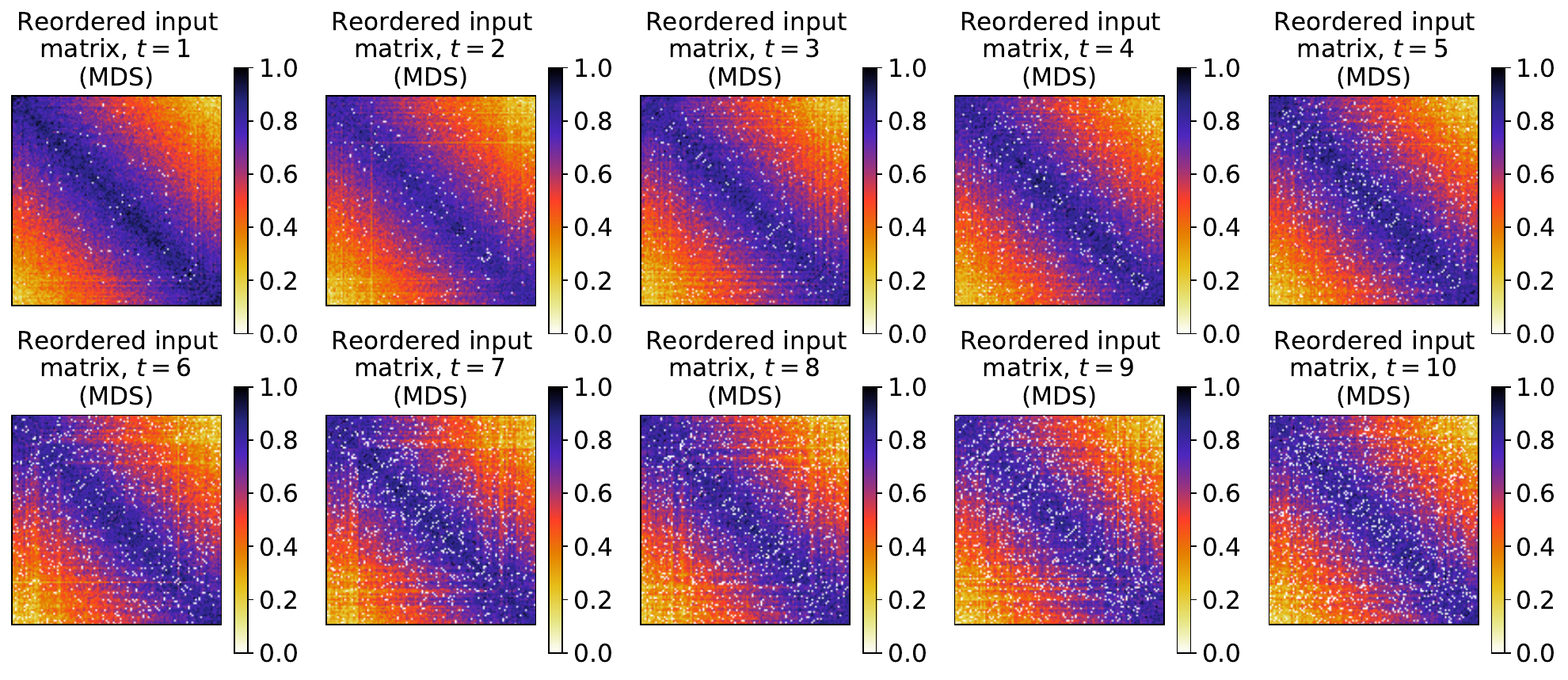}}
\caption{Examples of matrices $\underline{A}$ of MDS for different settings of $t$ (\textbf{undirected case with outliers}).}\vspace{3mm}
\label{fig:compare_u_underline_A_MDS_outlier}
\end{figure}

\begin{figure}[p]
\centerline{\includegraphics[width=0.9\hsize]{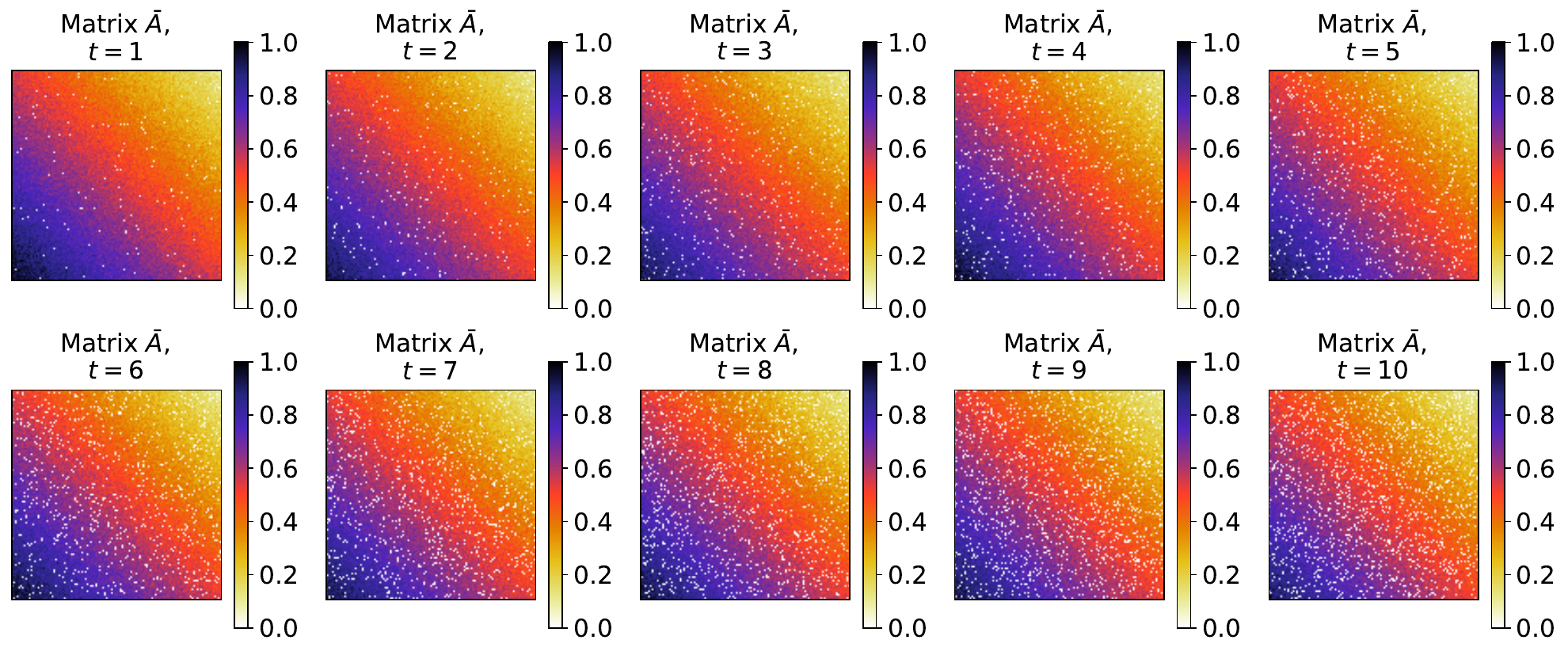}}
\caption{Examples of matrices $\bar{A}$ for different settings of $t$ (\textbf{directed case with outliers}).}\vspace{3mm}
\label{fig:compare_d_bar_A_outlier}
\centerline{\includegraphics[width=0.9\hsize]{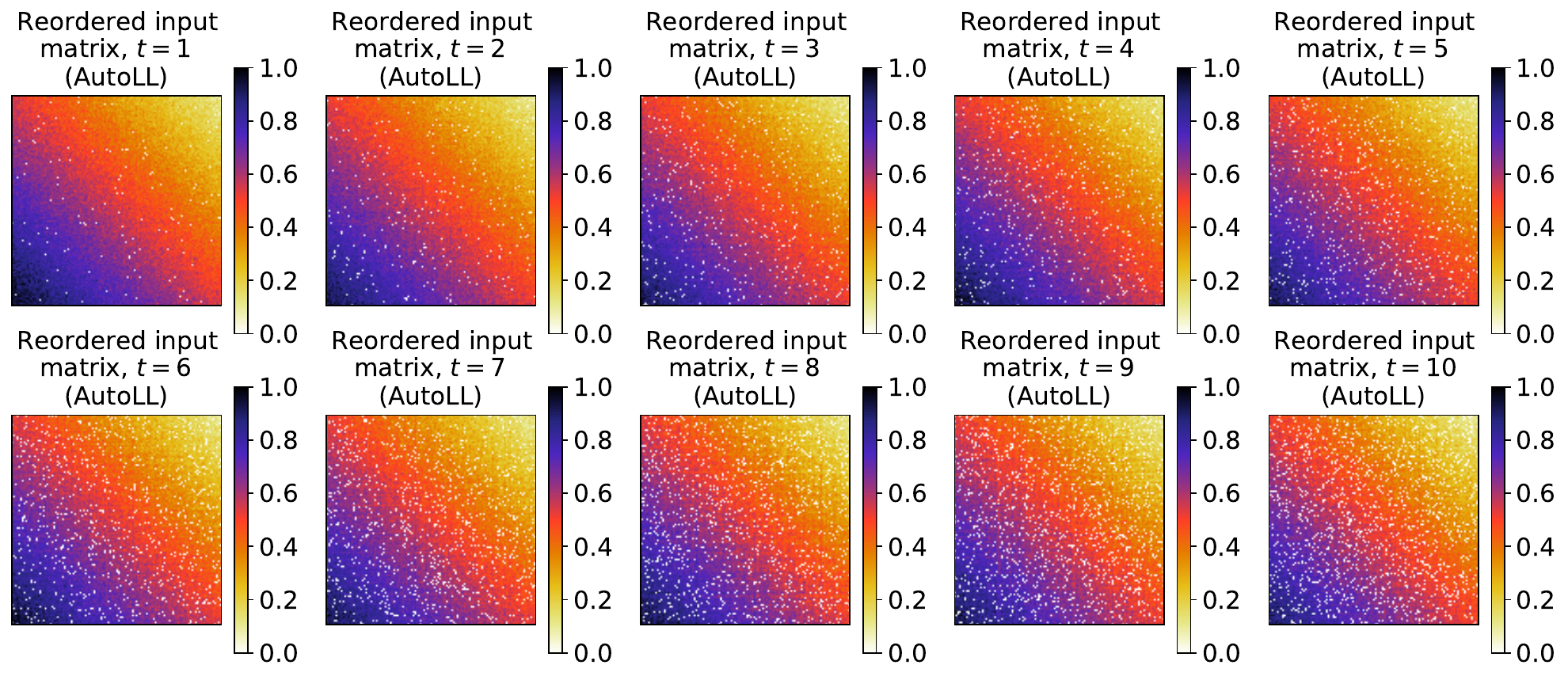}}
\caption{Examples of matrices $\underline{A}$ of the proposed AutoLL for different settings of $t$ (\textbf{directed case with outliers}).}\vspace{3mm}
\label{fig:compare_d_underline_A_AutoLL_outlier}
\centerline{\includegraphics[width=0.9\hsize]{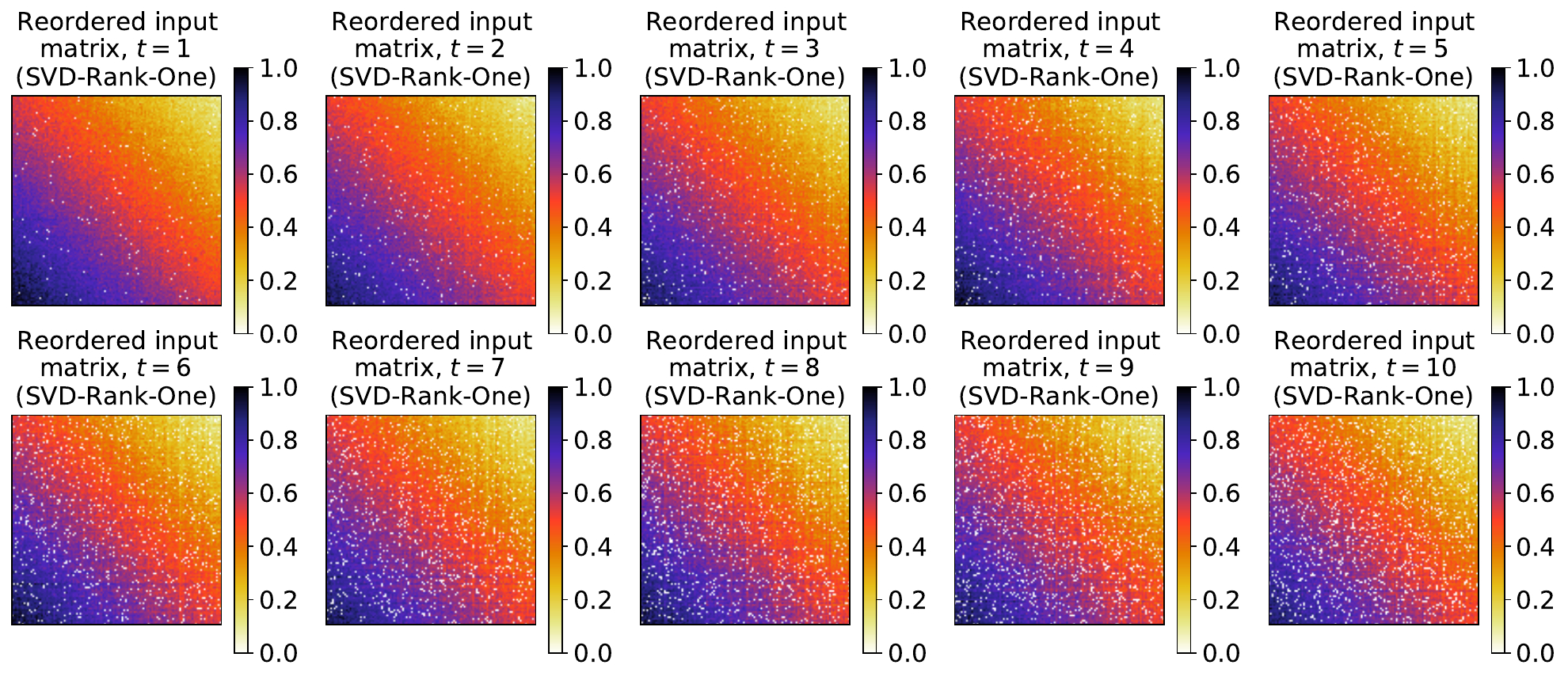}}
\caption{Examples of matrices $\underline{A}$ of SVD-Rank-One for different settings of $t$ (\textbf{directed case with outliers}).}\vspace{3mm}
\label{fig:compare_d_underline_A_SVD_Rank_One_outlier}
\end{figure}

\begin{figure}[t]
\centerline{\includegraphics[width=0.9\hsize]{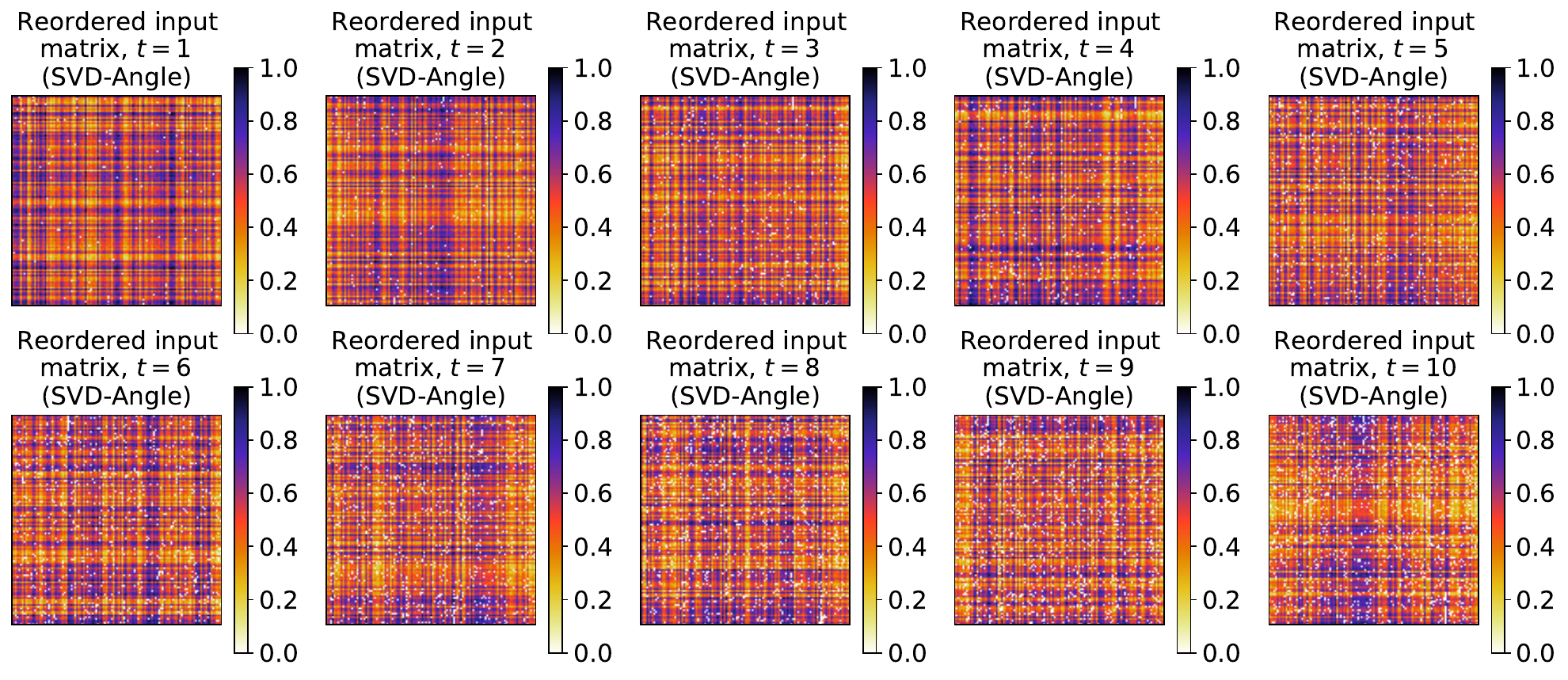}}
\caption{Examples of matrices $\underline{A}$ of SVD-Angle for different settings of $t$ (\textbf{directed case with outliers}).}\vspace{3mm}
\label{fig:compare_d_underline_A_SVD_Angle_outlier}
\centerline{\includegraphics[width=0.9\hsize]{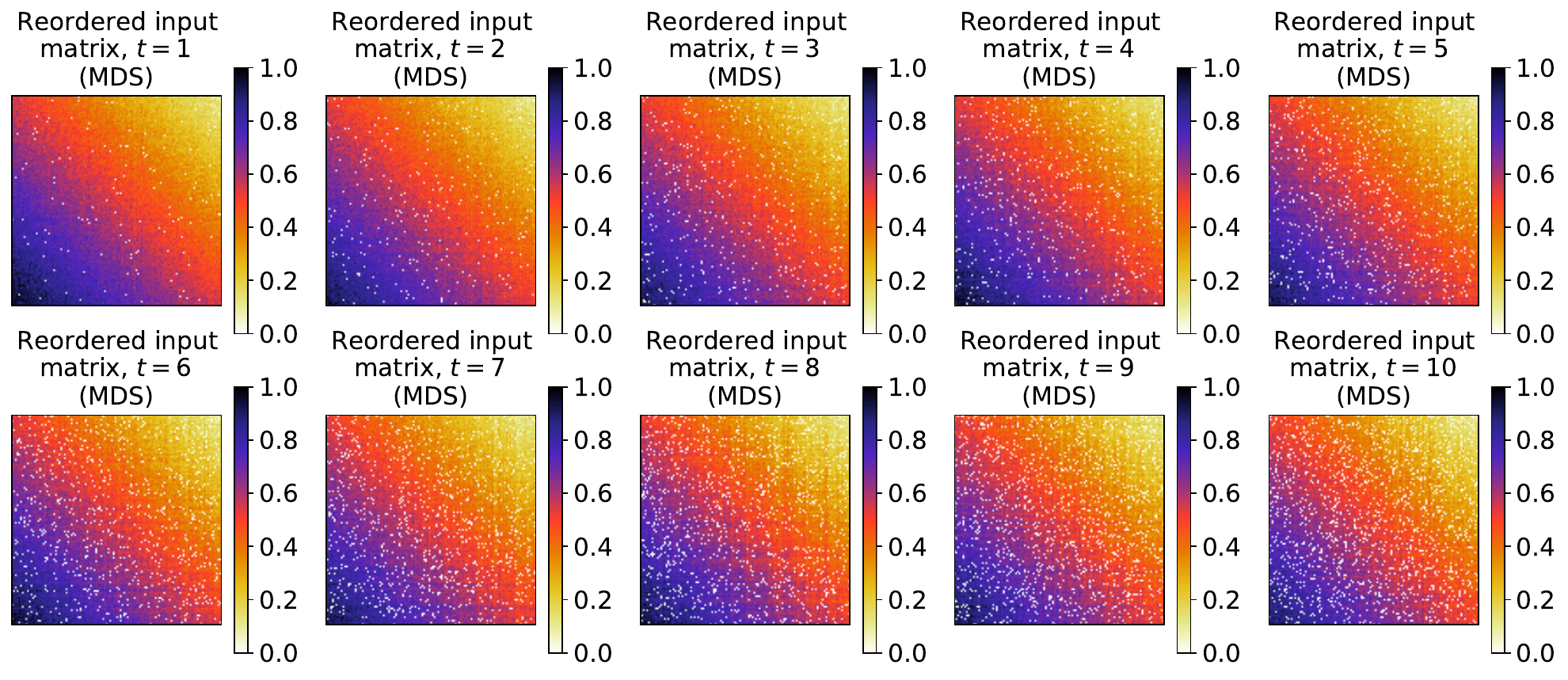}}
\caption{Examples of matrices $\underline{A}$ of MDS for different settings of $t$ (\textbf{directed case with outliers}).}\vspace{3mm}
\label{fig:compare_d_underline_A_MDS_outlier}
\end{figure}

\begin{figure}[t]
\centerline{\includegraphics[width=0.6\hsize]{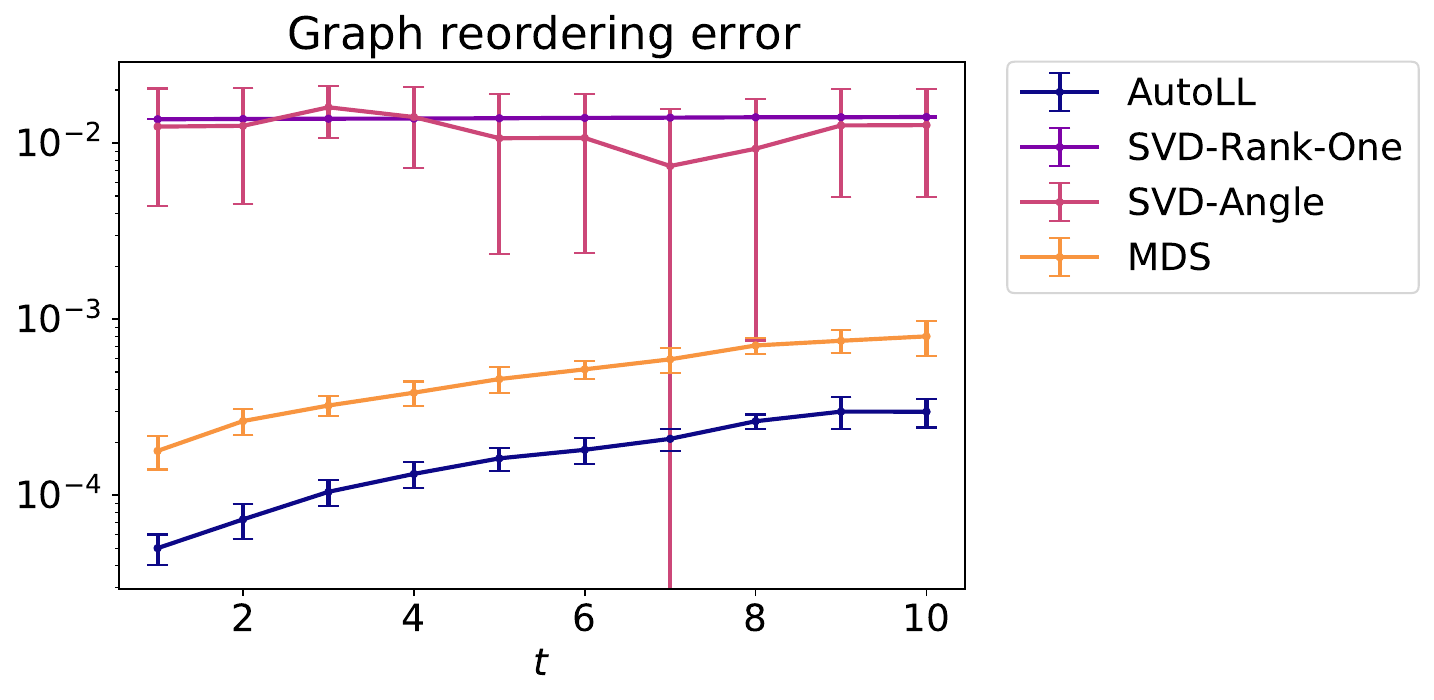}}
\caption{Average graph reordering errors for $10$ trials (\textbf{undirected case with outliers}). Error bars indicate sample standard deviation.}\vspace{3mm}
\label{fig:compare_u_outlier}
\centerline{\includegraphics[width=0.6\hsize]{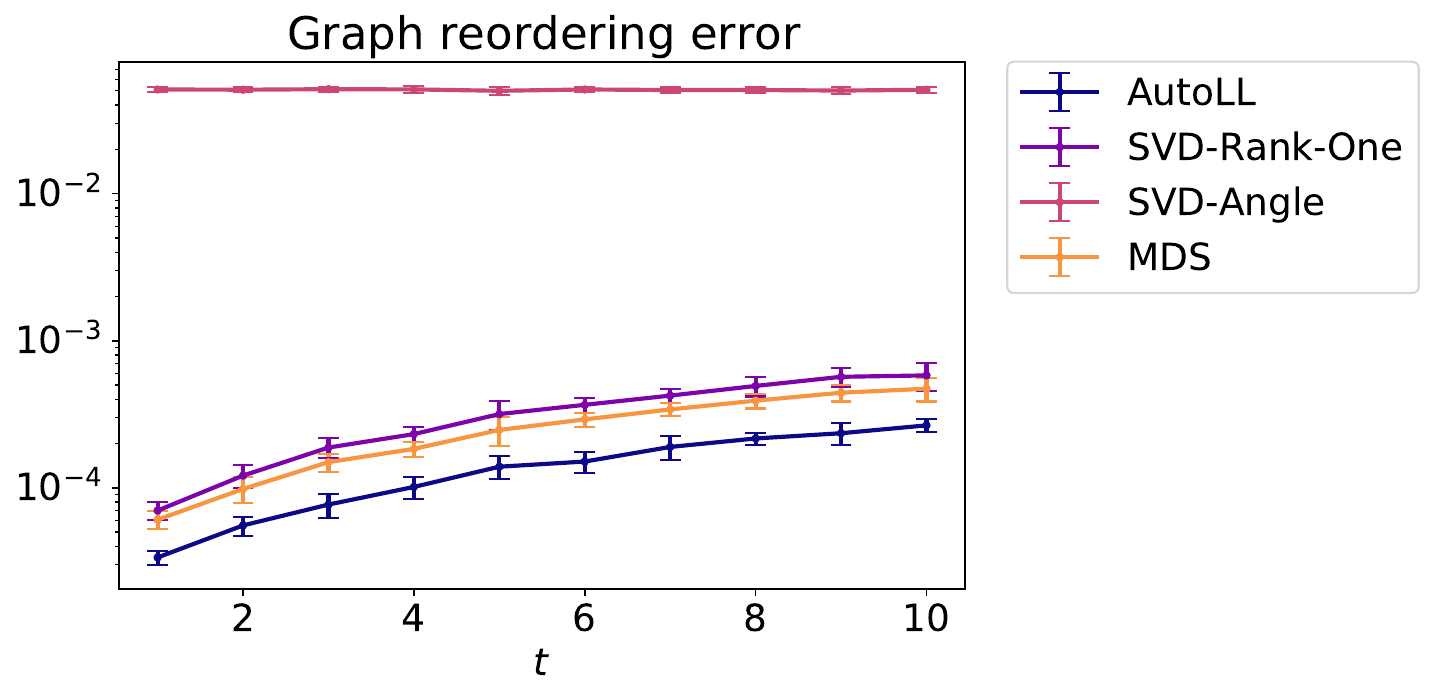}}
\caption{Average graph reordering errors for $10$ trials (\textbf{directed case with outliers}). Error bars indicate sample standard deviation.}
\label{fig:compare_d_outlier}
\end{figure}

Fig.~\ref{fig:compare_u_bar_A_outlier} (\ref{fig:compare_d_bar_A_outlier}) shows the examples of the original adjacency matrices $\bar{A}$ with the correct node orders, and Figs.~\ref{fig:compare_u_underline_A_AutoLL_outlier}, \ref{fig:compare_u_underline_A_SVD_Rank_One_outlier}, \ref{fig:compare_u_underline_A_SVD_Angle_outlier}, and \ref{fig:compare_u_underline_A_MDS_outlier} (\ref{fig:compare_d_underline_A_AutoLL_outlier}, \ref{fig:compare_d_underline_A_SVD_Rank_One_outlier}, \ref{fig:compare_d_underline_A_SVD_Angle_outlier}, and \ref{fig:compare_d_underline_A_MDS_outlier}), respectively, show those of the reordered observed matrices of the proposed AutoLL, SVD-Rank-One, SVD-Angle, and MDS in the undirected (directed) newtork settings. 
Figs.~\ref{fig:compare_u_outlier} and \ref{fig:compare_d_outlier}, respectively, show the graph reordering errors with different settings of $t$ in undirected and directed cases. From these figures, it may be observed that in both directed and undirected cases, the proposed AutoLL achieved the best average performance among the four methods. 


\subsection{Experiments using practical datasets}
\label{sec:exp_practical}

Finally, we applied the proposed AutoLL models to practical datasets and checked the reordering results. 

\paragraph{American college football network data} As an example of an undirected network, we used American college football network data \cite{Girvan2002}, whose nodes and edges represent the United States college football teams and the regular-season games between them, respectively. This network is unweighted (i.e., $A_{ij} \in \{0, 1\}$ for all $(i, j) \in \mathcal{J}$), and the number of nodes is $n = 115$. We trained the AutoLL-U model with this adjacency matrix and reordered it using the trained model. 

\begin{figure}[t]
\centerline{\includegraphics[width=0.33\hsize]{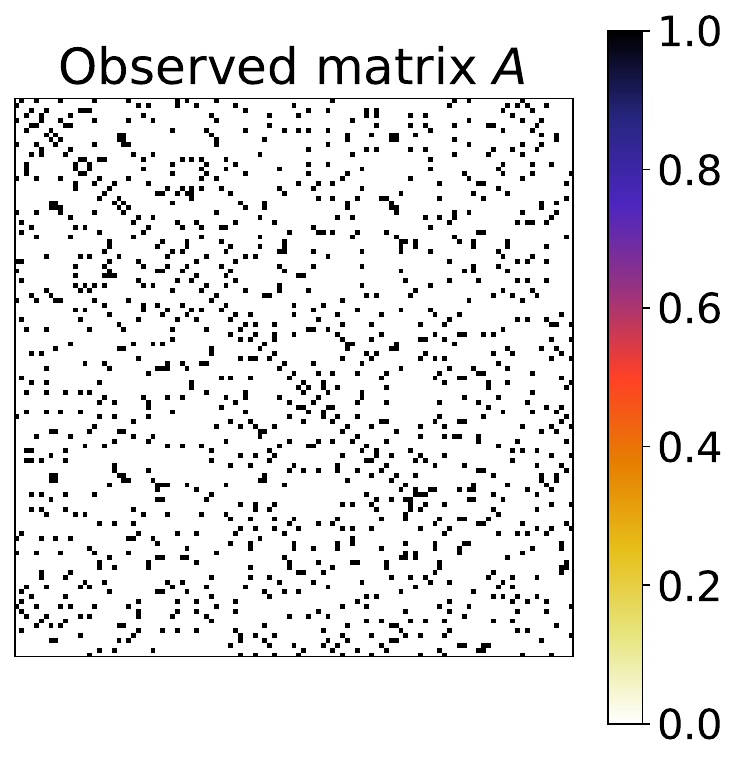}
\includegraphics[width=0.33\hsize]{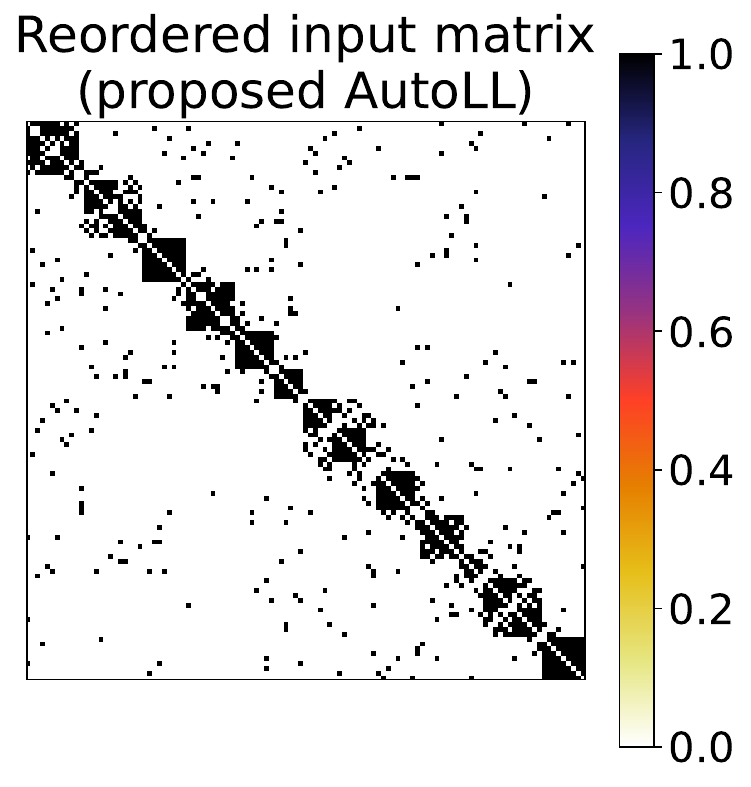}
\includegraphics[width=0.33\hsize]{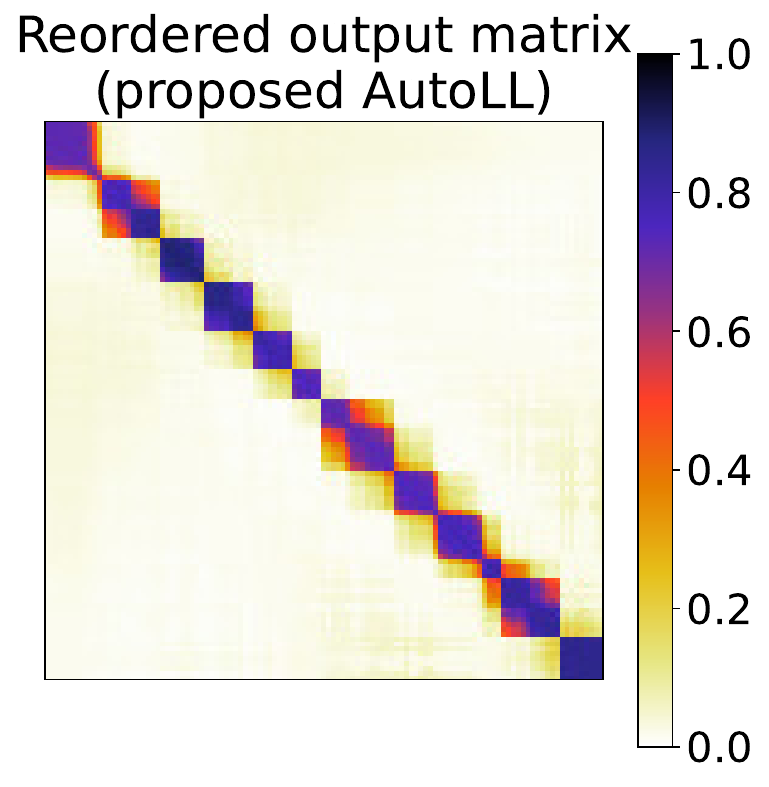}}
\centerline{\includegraphics[width=0.48\hsize]{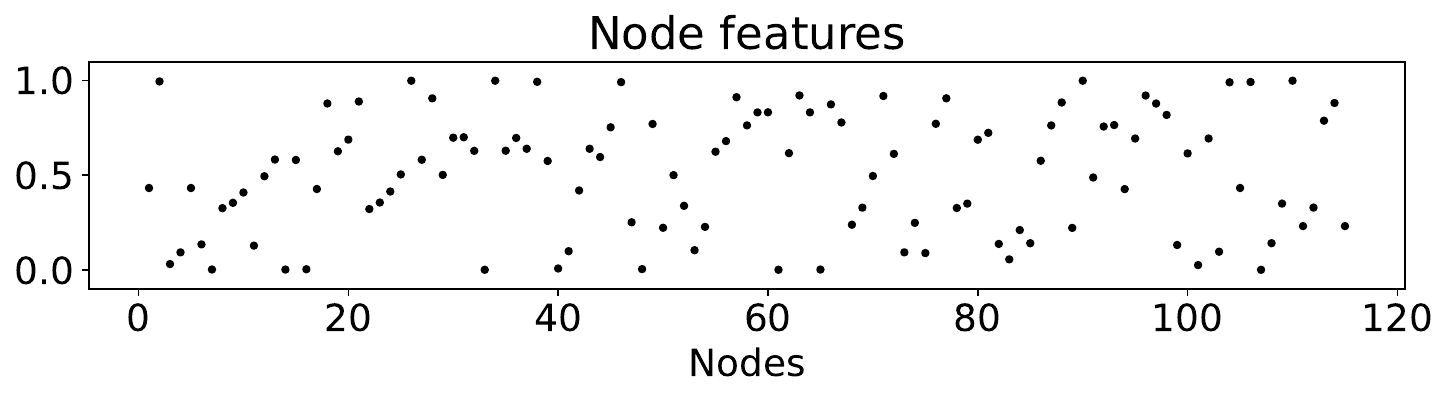}
\includegraphics[width=0.48\hsize]{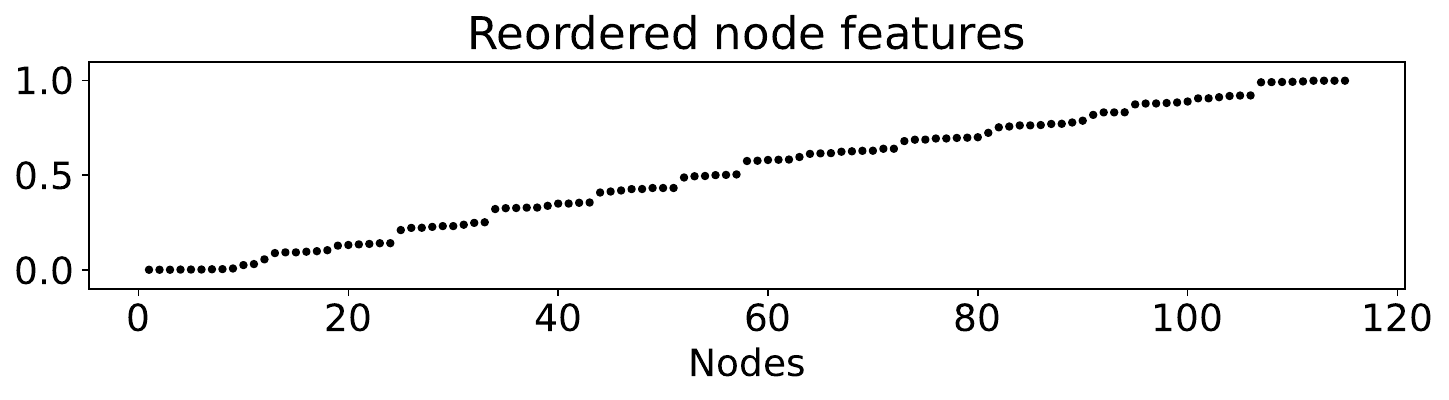}}
\caption{Matrices $A$, $\underline{A}$, and $\underline{\hat{A}}$ and vectors $\bm{z}$ and $\underline{\bm{z}}$ (\textbf{American college football network data}).}
\label{fig:practical_u}
\end{figure}

Fig.~\ref{fig:practical_u} shows the results of American college football network data. From the reordered observed and reconstructed matrices, $\underline{A}$ and $\underline{\hat{A}}$, respectively, it may be observed that this network had a community structure and the subgraph of each community was relatively dense. The reordered node feature $\underline{\bm{z}}$ also reflects such a community structure and it has stepped entry values. 

\paragraph{Neural network data of of C. elegans} We also applied the proposed AutoLL-D to the neural network data of C. elegans \cite{Watts1998, White1986}. The network was directed and weighted. We denote the original adjacency matrix of this dataset as $A^{(0)} = (A^{(0)}_{ij})_{1 \leq i, j \leq n}$, where $n = 297$. The weight $A^{(0)}_{ij}$ of the edge between the $i$th and $j$th nodes satisfies $A^{(0)}_{ij} \in \{ 0, 1, 2 \}$ for all $(i, j) \in \mathcal{J}$. There were $n_0 = 85,864$, $n_1 = 2,331$, and $n_2 = 14$ edges with weights of zero, one, and two, respectively. As in Section \ref{sec:exp_syn}, we defined the observed adjacency matrix $A$ by normalizing the original one $A^{(0)}$ such that all the entries of matrix $A$ were within the interval of $[0, 1]$. Because most of the edges had zero weights, training the proposed AutoLL-D model with the entire dataset tends to always output zero (i.e., $\underline{\hat{A}}_{ij} \approx 0$ for all $(i, j) \in \mathcal{J}$). To avoid this problem, we applied undersampling to the original data before training the AutoLL model. Specifically, we randomly selected $8 (n_1 + n_2)$ zero-weight edges without replacement and used them and all nonzero weight edges as a training dataset. We trained the AutoLL-D model with this dataset and reordered the entire adjacency matrix with the trained model. 

\begin{figure}[t]
\centerline{\includegraphics[width=0.33\hsize]{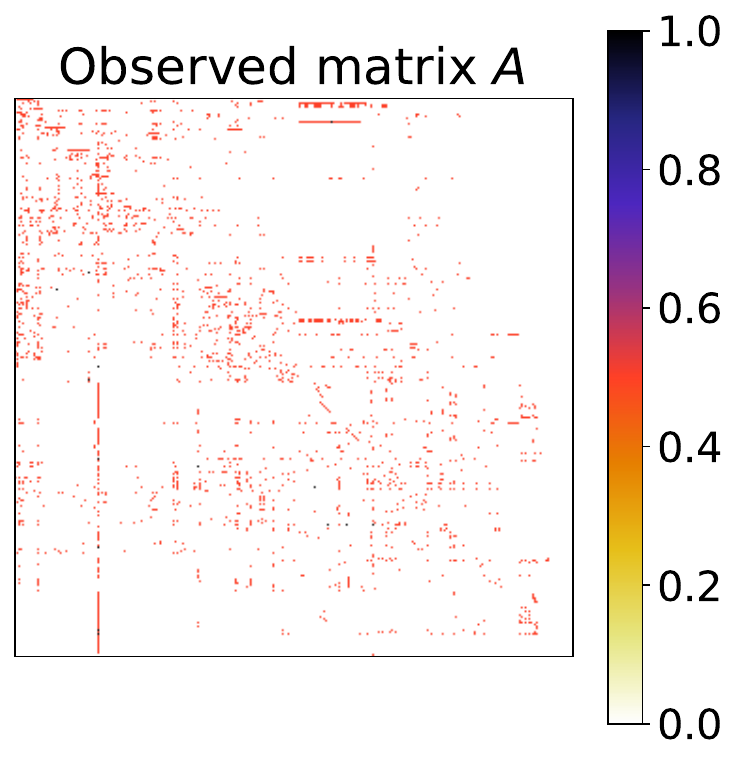}
\includegraphics[width=0.33\hsize]{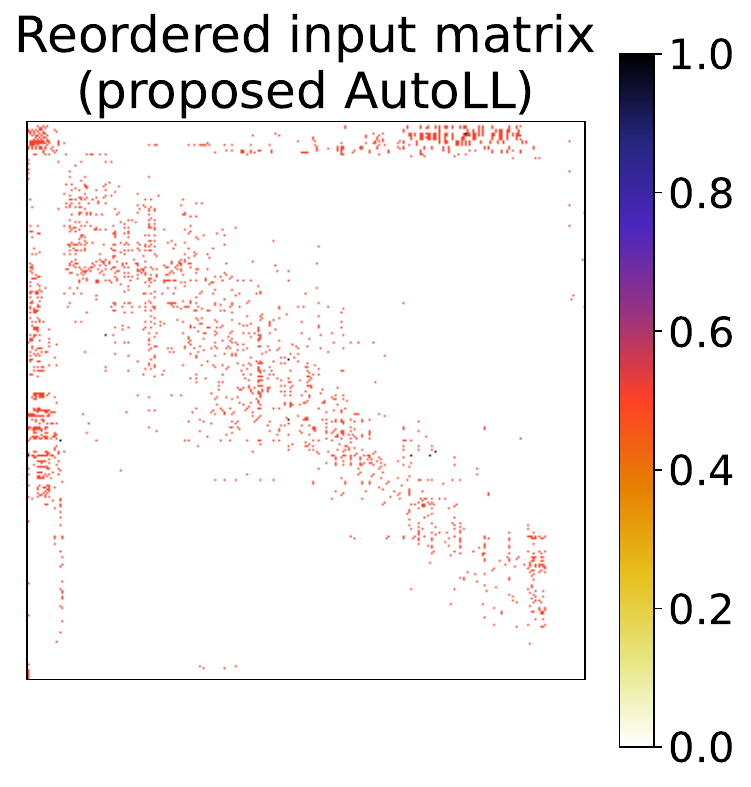}
\includegraphics[width=0.33\hsize]{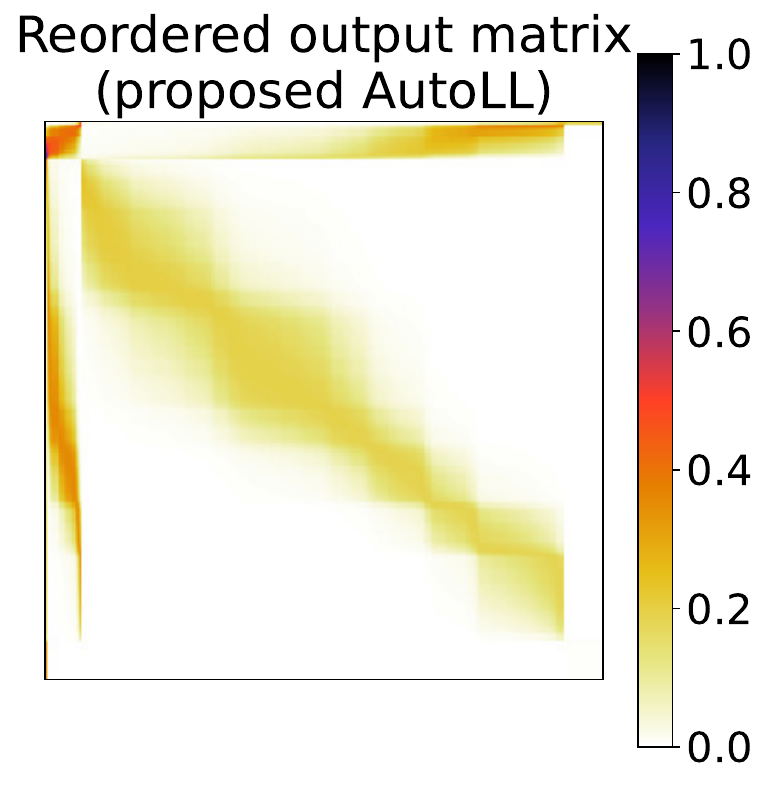}}
\centerline{\includegraphics[width=0.48\hsize]{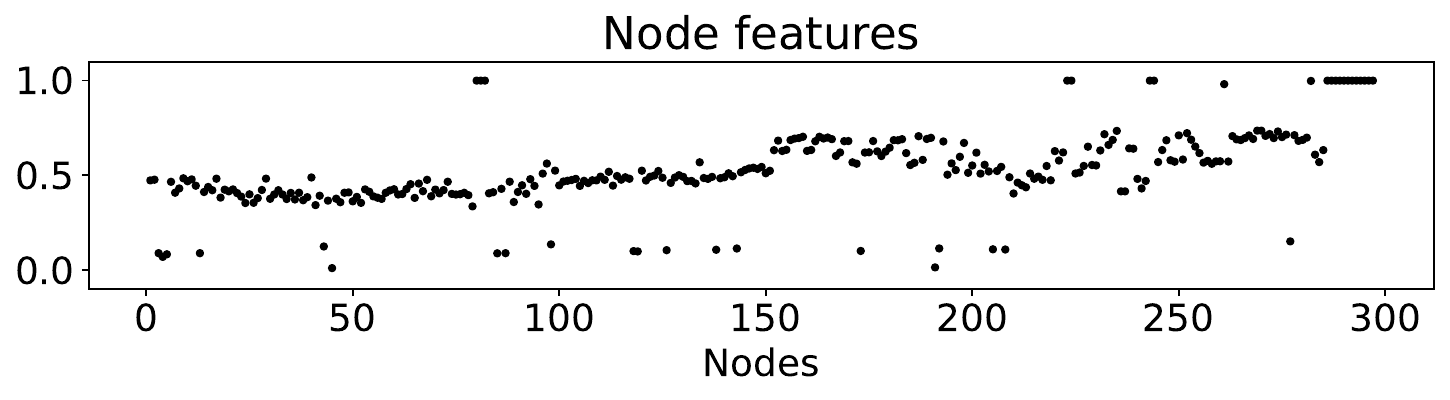}
\includegraphics[width=0.48\hsize]{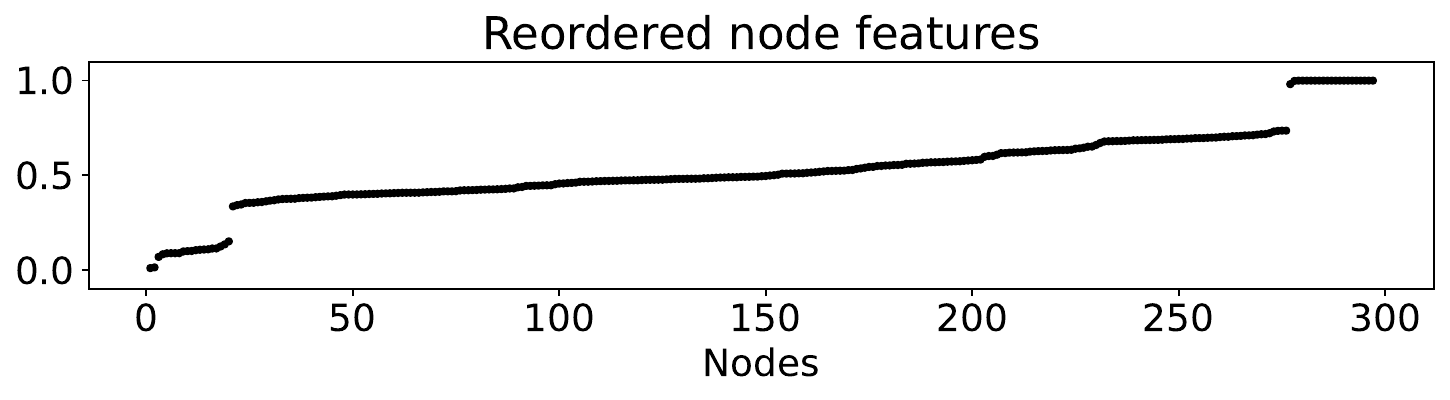}}
\caption{Matrices $A$, $\underline{A}$, and $\underline{\hat{A}}$ and vectors $\bm{z}$ and $\underline{\bm{z}}$ (\textbf{neural network data of C. elegans}).}
\label{fig:practical_d}
\end{figure}

Fig.~\ref{fig:practical_d} shows the results of the neural network data of C. elegans. From the reordered observed and reconstructed matrices, $\underline{A}$ and $\underline{\hat{A}}$, respectively, we see that there were several submatrices with different weight patterns. For instance, the average weights of the edges were relatively large in the upper left part of the reordered matrix $\underline{A}$. There was also a relatively large submatrix with a diagonal gradation pattern in the middle part of matrix $\underline{A}$. 

\section{Discussion}
\label{sec:discussion}

We discuss the proposed AutoLL in terms of the structural pattern of an observed adjacency matrix that appears by graph reordering and possible extensions of the problem setting. 

A notable advantage of the proposed AutoLL is its ability to automatically extract node features without assuming a specific structural pattern (e.g., block structure) in a given adjacency matrix. However, it would be more preferable if user-defined constraints could be freely added to the reordering result. It might be possible to induce a specific structural pattern in a reordered matrix by adding a penalty term to the training loss for the discrepancy from the target structure. 
Another feature of the proposed AutoLL is that it can provide a denoised mean adjacency matrix $\underline{\hat{A}}$, which shows the global structure of the reordered matrix. That is, we need another method to capture the finer structure of a given matrix. In terms of visualization, it would be interesting to simultaneously obtain the structural information of an observed adjacency matrix at multiple resolutions (e.g., structural patterns within $n_0 \times n_0$ submatrices for $n_0 = 10, 20, \dots$). 

There are several possible extended problem settings for graph reordering. For instance, tensor reordering has been considered in some recent studies \cite{Li2019, Smith2015}, where the goal is to reorder the indices of a $d$-way array ($d \geq 3$). 
A naive extension for such a case is to use $d$ input vectors to reconstruct each entry. However, because the model size increases with the input size, considerable computational and memory costs are required to train this naive model. 
Another possible problem setting is online learning using a temporal network. In this case, we obtain a sequence of snapshots of adjacency matrices $\{A_t\}$ for time steps $t = 1, 2, \dots$, and estimate the node order of matrix $A_{t+1}$ based on the model trained with matrices $A_1, \dots, A_t$. Instead of training the model with all the past matrices based on the same loss as in \eqref{eq:loss}, it would be more appropriate to fine-tune the current model only with the newest training data $A_t$ and to impose an additional constraint that the node orders do not change drastically between the two matrices of adjacent time steps. Future studies should provide more sophisticated and lightweight models for these extended settings. 

\section{Conclusions}
\label{sec:conclusion}

We developed a new neural-network-based graph reordering method, AutoLL, for both directed and undirected network data. By adopting an autoencoder-like architecture, the encoder part of the proposed AutoLL model automatically extracts the node features from a given observed adjacency matrix, which can be used to reorder the node indices. Moreover, an output matrix of a trained AutoLL model can be seen as a denoised mean adjacency matrix, which provides information on the global structural pattern in the observed adjacency matrix. We demonstrated the effectiveness of AutoLL experimentally through both qualitative and quantitative evaluations. 

\section*{Acknowledgment}

TS was partially supported by JSPS KAKENHI (18K19793, 18H03201, and 20H00576), Japan Digital Design, Fujitsu Laboratories Ltd., and JST CREST. 
We would like to thank Editage (\url{www.editage.com}) for English language editing.


\clearpage
\bibliographystyle{abbrv}
\bibliography{template}

\end{document}